\documentclass[11pt]{article}

\usepackage[preprint]{acl}

\usepackage{times}
\usepackage{latexsym}

\usepackage[T1]{fontenc}

\usepackage[utf8]{inputenc}

\usepackage{microtype}

\usepackage{inconsolata}

\usepackage{graphicx}

\usepackage{amsmath, amssymb, amsfonts, amsthm}
\usepackage{xcolor}
\usepackage{algorithm}
\usepackage{algpseudocode}

\usepackage{booktabs}
\usepackage{mathtools}
\usepackage{mathpartir}                %
\usepackage{stmaryrd}               %

\newcommand{\kw}[1]{\mathsf{#1}}                  %
\newcommand{\srt}[1]{\mathsf{#1}}                 %

\newcommand{\Src}{\mathrm{Src}}
\newcommand{\Tst}{\mathrm{Test}}
\newcommand{\Cseed}{C_{\text{seed}}}
\newcommand{\Fgold}{\mathcal{F}_{\text{gold}}}
\newcommand{\Tkw}{\mathcal{T}_{\text{kw}}}

\newcommand{\Kmap}{\mathcal{K}}

\newcommand{\denote}[1]{\llbracket #1 \rrbracket}
\newcommand{\cand}{\kw{cand}}
\newcommand{\rw}{\kw{rewrite}}
\newcommand{\bind}{\mathbin{\gg\!=}}

\newcommand{\Gnat}{g_{\text{unt}}}
\newcommand{\Gid}{g_{\text{id}}}

\newtheorem{proposition}{Proposition}

\newtheorem{definition}{Definition}

\title{Programmable Distributions of Semantics-Preserving Variants for Evaluating Code Agents}

\author{
  \textbf{Hasan Najib Mahmud}\textsuperscript{1}\thanks{Equal contribution.},
  \textbf{Shreya Gupta}\textsuperscript{2}\footnotemark[1],
  \textbf{Isha Chaudhary}\textsuperscript{3},
  \textbf{Nathaniel Enis}\textsuperscript{1},
  \\
  \textbf{Ravi Mangal}\textsuperscript{1},
  \textbf{Gagandeep Singh}\textsuperscript{3},
  \textbf{Corina Pasareanu}\textsuperscript{4}
  \\
  \textsuperscript{1}Colorado State University,
  \textsuperscript{2}Microsoft,
  \\
  \textsuperscript{3}University of Illinois Urbana-Champaign,
  \textsuperscript{4}Carnegie Mellon University
  \\
  \small{\textbf{Correspondence:} \href{mailto:Hasannajib.Mahmud@colostate.edu}{Hasannajib.Mahmud@colostate.edu}}
}

\begin{document}
\maketitle
\begin{abstract}
Code agents are increasingly deployed to resolve real software issues, yet their reliability under superficial code variations remains poorly understood. We evaluate whether coding agents that repair repository-level issues remain reliable when the surrounding codebase is rewritten into a semantically equivalent form. We sample semantically equivalent variants of a repository from an explicitly defined distribution and compare the agent's resolve rate on them with that on the original. Different distributions reveal different aspects of an agent, such as its sensitivity to routine refactoring or to misleading code near the location of a fix. We develop VDL, a probabilistic programming language for expressing such distributions, whose programs randomly apply semantics-preserving transformations (SPTs) to a repository. For our evaluation, we use it to define a generic distribution that is not targeted at any particular failure mode. We evaluate two agentic scaffolds (mini-SWE agent and OpenCode), each backed by one of four frontier models (Claude Opus 4.5, Kimi K2.5, MiniMax M2.5, and Qwen 3.6-27B), on instances from SWE-bench Verified and SWE-bench Pro. Resolve rates drop by up to 6.7 percentage points, significantly in 6 of 16 configurations. No ranking of models by robustness holds across scaffolds, revealing a jagged robustness frontier.

\end{abstract} 

\noindent\textbf{Code:} \url{https://github.com/CSU-TrustLab/jagged-frontier}

\section{Introduction}
\label{sec:introduction}

In the span of a few years, AI-powered coding tools have moved from research prototypes into mainstream software development, with developer surveys reporting that a majority of professionals now use them regularly~\cite{stackoverflow2025, github2025octoverse, jetbrains2025, dora2025}.
Their progress is charted on challenging coding benchmarks such as SWE-bench Verified~\cite{openai2024swebenchverified} and SWE-bench Pro~\cite{deng2025swe}, against which nearly every new model and agent is measured. While performance numbers on these benchmarks can be instructive, they do not account for the fact that neural models, in general, are known to learn shortcuts~\cite{geirhos2020shortcut} and therefore be susceptible to small input perturbations. Two concerns follow. First, if agent behavior shifts with superficial changes to the surrounding code, then benchmark numbers may overstate deployment reliability. Second, such shifts would suggest that the underlying models rely on shallow syntactic patterns rather than an understanding of program semantics. Prior work has evaluated the robustness of code LLMs to semantics-preserving perturbations in single-turn settings~\cite{yefet2020adversarial, ramakrishnan2022semantic, wang2023recode}. However, the robustness of repository-level code agents, which interact with the LLM over many turns, has received little systematic study. We present, to our knowledge, the first such evaluation, and find a \emph{jagged robustness frontier}: a model robust under one scaffold or codebase can be the most brittle under another, so benchmark rankings do not transfer to deployment.

\paragraph{Approach.}
For robustness evaluation, we compare agent performance on unperturbed code repositories with \emph{semantically equivalent} perturbed variants. 
The perturbations are injected via local, semantics-preserving transformations (SPTs) of the code. We develop a library of SPTs that either mimic refactorings a developer may routinely perform or are intended to stress-test the agent. We draw variants from an explicitly defined probability distribution, which specifies which SPTs are applied, in which files, and at how many locations. The choice of distribution determines which aspect of an agent the evaluation tests: rewriting only the files that the agent must edit to resolve an issue probes something different from rewriting only the rest of the repository. To express such distributions, we develop VDL (Variant Distribution Language), a small probabilistic programming language. A VDL program applies SPTs to a repository, making random choices about which SPTs to apply and where, so that each run of the program yields a different variant. Every well-formed program yields only variants that behave like the original repository and are checked by the same tests.

\paragraph{Untargeted generator.}
We use VDL to define the distribution for our evaluation, a program we call the \emph{untargeted generator}. It randomly chooses which SPTs to apply and where to apply them, drawing from the full catalog without targeting any particular failure mode. Each variant is constructed independently, and unlike algorithms that aim to find adversarial examples~\cite{srikant2021generating}, the sampling process is not guided by feedback computed from model outcomes. This non-adversarial design is intentional: our aim is to compute \emph{lower bounds} on the impact of SPTs on code agents, since a feedback-guided adversary can only do more damage. Such bounds can inform which code agents to deploy and guide the design of more robust models and scaffolds.

\paragraph{Stochastic agents.}
The stochastic nature of LLM-based agents makes the robustness evaluation challenging.
If an agent has a failing run on a perturbed variant but a passing run on the unperturbed repository, this observation is not sufficient to conclude that the perturbation was the cause; the observation might simply be an artifact of agent stochasticity. Our methodology therefore isolates the perturbation effect from run-to-run variability while keeping the number of costly agent runs manageable. We run the agent repeatedly on both the seed and its variants and report confidence intervals for the resulting change in resolve rate.

\paragraph{Experiments and findings.}
We evaluate two agentic scaffolds (mini-SWE agent and OpenCode), each backed by one of four frontier models (Claude Opus 4.5, Kimi K2.5, MiniMax M2.5, and Qwen 3.6-27B), on 54 repository-level issue-resolution instances drawn from SWE-bench Verified and SWE-bench Pro. Our experiments reveal the following insights: (1) localized perturbations applied without feedback cause small degradations in most configurations, up to 6.7 percentage points and statistically significant in 6 of 16 configurations; (2) perturbations raise the effort an agent expends, increasing step counts and token cost by up to 9.9\% and 23.0\% even where resolve rates barely change; (3) within a scaffold, more capable models such as Claude Opus 4.5 can be affected more than weaker ones, and across scaffolds the simpler one tends to be more robust; (4) the scaffold and the benchmark change the robustness ranking of models: Qwen is among the most robust under mini-SWE agent on SWE-bench Verified yet the most brittle under OpenCode; and (5) the effect concentrates in a small set of instances. Together these observations indicate that robustness is a joint property of the model, the scaffold, and the workload, not of the model alone, revealing a \emph{jagged robustness frontier}.

\section{VDL: A Language for Variant Distributions}
\label{sec:lang}

This section presents our probabilistic programming language, VDL, for sampling variants of a repository, shows that its programs generate only variants that pose the same task as the seed, and defines the quantities our experiments estimate.

\subsection{Tasks, Equivalence, and Transformations}
\label{subsec:lang-prims}

\paragraph{Task instances.}
Throughout the paper, lowercase letters denote single values, uppercase letters denote sets and random variables, and calligraphic letters denote fixed sets given by the catalog or the instance; counts ($N$, $K$, $R$) and distributions ($D$) keep their usual uppercase. A repository $C$ is a collection of files, and so is written in uppercase. We write $\Src(C)$ for its source files and $\Tst(C)$ for its test files; a repository may also contain other files, such as documentation. A \emph{task instance} $i$ consists of a seed repository $\Cseed$ at a base commit, an issue description, and two sets of test cases, defined in the test files $\Tst(\Cseed)$, that a patch must satisfy: the \texttt{FAIL\_TO\_PASS} tests fail on $\Cseed$ and must pass once the issue is resolved, and the \texttt{PASS\_TO\_PASS} tests pass on $\Cseed$ and must keep passing. Each instance also has a \emph{gold patch}, a reference edit resolving the issue; we write $\Fgold \subseteq \Src(\Cseed)$ for the files it modifies. For a repository $C$, $i[C]$ denotes instance $i$ with its seed repository replaced by $C$.

\paragraph{Observational equivalence.}
Two repositories $C$ and $C'$ are \emph{observationally equivalent}, written $C \equiv C'$, if their code produces the same observable behavior on every possible input. Concretely, under any input, $C$ and $C'$ must (i)~return the same value or raise the same exception, and (ii)~produce the same externally observable effects. If either halts, so must the other. The relation $\equiv$ is transitive.

\paragraph{Transformations.}
A transformation $t$ rewrites code at the locations in a file where its structural pattern matches, which we call its \emph{candidate sites}. We write $\cand(C, t, f, k)$ for the candidate sites of $t$ in file $f$ of $C$, and $\rw(C, t, f, S, k)$ for the repository obtained by rewriting $f$ at the sites $S \subseteq \cand(C, t, f, k)$; every other file is unchanged, and rewriting no sites leaves $C$ unchanged. The argument $k$ is a keyword. Some transformations, which we call \emph{keyword-bound}, insert code containing a given keyword, such as a dead method with a given name; their candidate sites can depend on the keyword, since, e.g., a class already defining a method with that name is skipped. The other transformations ignore $k$, and we pass them the placeholder $\bot$. Keywords are extracted from the issue description (Section~\ref{sec:sampler}); $\Kmap(t)$ holds the keywords available to $t$, empty unless $t$ is keyword-bound.

\begin{definition}
\label{def:spt}
A transformation $t$ is \emph{semantics-preserving} (an SPT) if $\rw(C, t, f, S, k) \equiv C$ for every repository $C$, source file $f \in \Src(C)$, keyword $k$, and set of candidate sites $S \subseteq \cand(C, t, f, k)$.
\end{definition}

\noindent VDL is parameterized by a catalog $\mathcal{T}$ of SPTs, of which the subset $\Tkw$ is keyword-bound. Section~\ref{sec:spt} describes our catalog of 14 and the differential testing that supports their being semantics-preserving. The evidence is empirical, so the guarantee in Section~\ref{subsec:lang-props} holds relative to it.

\begin{figure*}[t]
\small
\begin{align*}
F \sim \kw{scan}(p, n) &\triangleq F := \emptyset \mathbin{;} c := 0 \mathbin{;} \kw{for}\ f \in \Src(\Cseed)\ \kw{do} \\
&\qquad \big(u \sim \kw{unif}(0, 1) \mathbin{;} \kw{if}\ f \in \Fgold \lor (u \le p \land c < n)\ \kw{then}\ (F := F \cup \{f\} \mathbin{;} c := c + 1)\big) \\
W \sim \kw{targets}(t, n) &\triangleq W \sim \kw{choose}(n, \Kmap(t)) \mathbin{;} \kw{if}\ W = \emptyset\ \kw{then}\ W := \{\bot\} \\
\kw{edit}(t, k, F, \phi) &\triangleq \kw{for}\ f \in F\ \kw{do}\ \big(S := \cand(t, f, k) \mathbin{;} S' \sim \kw{choose}(\lceil \phi \cdot |S| \rceil, S) \mathbin{;} \rw(t, f, S', k)\big) \\
\noalign{\vskip 3pt\hrule\vskip 3pt}
\Gnat &\triangleq T_{\text{sel}} \sim \kw{choose}(N_t, \mathcal{T}) \mathbin{;} \kw{for}\ t \in T_{\text{sel}}\ \kw{do}\ \Big(p \sim \kw{unif}(0,1) \mathbin{;} \\
&\quad \kw{if}\ t \in \Tkw\ \kw{then}\ \big(W \sim \kw{targets}(t, N_k) \mathbin{;} \kw{for}\ k \in W\ \kw{do}\ (F \sim \kw{scan}(p, N_f) \mathbin{;} \kw{edit}(t, k, F, \phi))\big) \\
&\quad \kw{else}\ \big(F \sim \kw{scan}(p, \infty) \mathbin{;} \kw{edit}(t, \bot, F, \phi)\big)\Big)
\end{align*}
\caption{Top: syntactic sugar used in our generators, defined in terms of the basic VDL constructs; $\kw{if}\ b\ \kw{then}\ g$ abbreviates $\kw{if}\ b\ \kw{then}\ g\ \kw{else}\ \kw{skip}$, and the variables $c$, $u$, $f$, $S$, and $S'$ are fresh. Bottom: the untargeted generator $\Gnat$ used in our experiments (Section~\ref{sec:sampler}).}
\label{fig:sugar}
\label{fig:galg}
\end{figure*}

\subsection{Generators}
\label{subsec:lang-syntax}

A VDL program, which we call a \emph{generator} after the random input generators of property-based testing~\cite{claessen2000quickcheck}, takes a repository and returns a randomly transformed copy. VDL is a small imperative probabilistic programming language: besides assignments, conditionals, and loops, its programs can draw values from probability distributions. Its basic constructs are expressions $e$, distributions $d$, and generators $g$:
\begin{align*}
e &::= x \mid c \mid \mathit{op}(e, \dots, e) \mid \Kmap(t) \mid \cand(t, f, k) \\
d &::= \kw{choose}(n, s) \mid \kw{unif}(n, n) \\
g &::= \kw{skip} \mid g \mathbin{;} g \mid x := e \mid x \sim d \\
  &\quad\ \mid \rw(t, f, s, k) \mid \kw{if}\ b\ \kw{then}\ g\ \kw{else}\ g \\
  &\quad\ \mid \kw{for}\ x \in s\ \kw{do}\ g
\end{align*}
Here $n$, $b$, $s$, $t$, $f$, and $k$ stand for expressions denoting a number, a Boolean, a finite set, a transformation, a file, and a keyword, respectively.

\emph{Expressions} are built from variables $x$; constants $c$, namely transformations, keywords, $\bot$, numbers, and the sets $\mathcal{T}$, $\Tkw$, $\Src(\Cseed)$, and $\Fgold$; arithmetic ($+$, $\cdot$, $\lceil\cdot\rceil$), comparisons ($<$, $\le$, $=$), Boolean connectives ($\land$, $\lor$), and set operations ($\in$, $\cup$, $\setminus$, $|\cdot|$); and the two instance-specific functions $\Kmap(t)$ and $\cand(t, f, k)$, the latter giving the candidate sites of $t$ in file $f$ of the current repository. \emph{Distributions}: $\kw{choose}(n, X)$ draws $n$ elements of the finite set $X$ uniformly without replacement, or all of $X$ if it has fewer than $n$, and $\kw{unif}(a, b)$ draws a number uniformly from $[a, b]$. \emph{Generators} run on a repository and a store of variable values. $\kw{skip}$ does nothing, $g_1 \mathbin{;} g_2$ runs $g_1$ and then $g_2$, $x := e$ assigns the value of $e$ to $x$, and $x \sim d$ draws a value from $d$ and assigns it to $x$. Conditionals are standard, and $\kw{for}\ x \in X\ \kw{do}\ g$ runs $g$ once for each element of $X$, visiting the elements in a fixed order (catalog order for transformations, file order for files). The only construct that changes the repository is $\rw(t, f, S, k)$: if $f$ is a source file, it rewrites $f$ with $t$ and keyword $k$ at those sites in $S$ that are candidate sites, and otherwise it does nothing. Formally, a generator maps a state, consisting of a repository and a store, to a probability distribution over states; the store is part of the outcome because drawn values are assigned to variables (Appendix~\ref{app:lang}). Every well-formed program (Definition~\ref{def:wf}) therefore has a meaning.

\begin{definition}
\label{def:wf}
A generator is \emph{well formed} if it is well typed: each variable holds values of a single type throughout the program, and every construct receives arguments of the types it expects; for instance, the condition of an $\kw{if}$ must be a Boolean and the range of a $\kw{for}$ must be a finite set. Appendix~\ref{app:lang} gives the typing rules.
\end{definition}

\paragraph{Syntactic sugar.}
To keep generators short, Figure~\ref{fig:sugar} defines three constructs as \emph{syntactic sugar}: each is shorthand for a sequence of basic constructs. $F \sim \kw{scan}(p, n)$ draws a set of source files: gold-patch files are always included, and every other file is included with probability $p$ while fewer than $n$ files have been included. $W \sim \kw{targets}(t, n)$ draws up to $n$ keywords from $\Kmap(t)$, or returns $\{\bot\}$ if there are none. $\kw{edit}(t, k, F, \phi)$ rewrites a uniformly chosen fraction $\phi$ of the candidate sites of $t$ in each file of $F$. Figure~\ref{fig:galg} (bottom) shows the generator used in our experiments, which Section~\ref{sec:sampler} describes.

For a well-formed generator $g$ and an instance $i$, the \emph{variant distribution} $D_i(g)$ is the distribution of the repository component of the state obtained by running $g$ on the seed repository of $i$, with the constants $\Fgold$ and $\Kmap$ taken from $i$.

\subsection{Generated Variants Pose the Same Task}
\label{subsec:lang-props}

\begin{proposition}
\label{prop:valid}
Let $g$ be a well-formed generator, $i$ an instance with seed repository $\Cseed$, and $C'$ a repository that $g$ produces from $\Cseed$ with positive probability. Then (a)~$C'$ has the same files and the same test files as $\Cseed$, and (b)~$C' \equiv \Cseed$.
\end{proposition}

\noindent Part (a) holds because $\rw$ changes only the contents of source files. Part (b) holds because every transformation in $\mathcal{T}$ is semantics-preserving and $\equiv$ is transitive. Appendix~\ref{app:lang} gives the proof. Together, the two parts mean that each variant poses the same task as the seed: its code behaves the same, and success is judged by the same \texttt{FAIL\_TO\_PASS} and \texttt{PASS\_TO\_PASS} tests. A change in how often an agent succeeds on the variants can therefore be attributed to how the code is written, and not to a change in the task.

\subsection{Measuring Robustness}
\label{subsec:lang-robustness}

Fix an agent $\alpha$: a scaffold, such as mini-SWE agent, together with the LLM that drives it. Agent runs are random, so we measure success by the probability $p(i)$ that a run on instance $i$ \emph{resolves} it, that is, produces a patch that passes both test sets; we leave $\alpha$ implicit in the notation. Averaging over the variants of a generator $g$ gives
\[
\mu_i(g) = \mathbb{E}_{C \sim D_i(g)}\big[p(i[C])\big],
\]
and the \emph{degradation} $\Delta_i(g) = \mu_i(\Gid) - \mu_i(g)$, our main quantity, compares $g$ with the identity generator $\Gid = \kw{skip}$: $\alpha$ is robust to the variation that $g$ describes when $\Delta_i(g)$ is near zero.

We estimate $\mu_i(g)$ and $\Delta_i(g)$ from sampled runs, as in statistical model checking~\cite{legay2010statistical} (Section~\ref{sec:experiments}). Because the variants are independent draws from an explicit distribution, and each run of $\alpha$ on a variant is an independent trial, these estimates come with statistical guarantees: standard concentration bounds give intervals that contain $\mu_i(g)$ and $\Delta_i(g)$ with a stated probability (Appendix~\ref{subsec:est_guarantees}).

\section{The Untargeted Generator}
\label{sec:sampler}

The untargeted generator $\Gnat$ (Figure~\ref{fig:galg}, bottom) is the distribution used in our experiments. It is well formed, so by Proposition~\ref{prop:valid} every variant it produces poses the same task as the seed; Algorithm~\ref{alg:sampler} (Appendix~\ref{app:sampler_details}) gives it as pseudocode. It draws from the full catalog, targets no particular failure mode, and gets no feedback from the agent, so the degradation it measures is an average over its variants; an adversary that picks the variants where the agent does worst can only cause more degradation than $\Gnat$.

$\Gnat$ first draws $N_t = 3$ transformations uniformly from the catalog, so a variant typically combines rewrites that mimic developer refactorings with ones intended to stress-test the agent. For each selected transformation $t$, it draws an inclusion probability $p \sim \kw{unif}(0,1)$; $\kw{scan}$ then includes every gold-patch file and each other source file with probability $p$, so variants range from a localized edit to repository-wide perturbation. In each included file, $\kw{edit}$ rewrites a uniformly chosen fraction $\phi = 0.7$ of the candidate sites. For the keyword-bound transformations, we first prompt the LLM that backs the agent under evaluation to list the method names and string literals mentioned in the issue description; a keyword-bound $t$ is bound to up to $N_k = 5$ of these, drawn uniformly (method names for Dead Method Injection, string literals for Dead String Assignment). It then makes one pass over the files per keyword, all with the same $p$, and each pass includes at most $N_f = 10$ files, counting gold-patch files. Without this cap, agents frequently detected the perturbation and reverted the repository outright (for instance via \texttt{git reset}). These hyperparameter values were fixed in advance rather than tuned (Appendix~\ref{subsec:hyperparams}). Appendix~\ref{app:alt_generators} gives alternatives to $\Gnat$.

\section{Semantics-Preserving Transformations}
\label{sec:spt}
Our catalog contains 14 SPTs: If Else Switcher, For Loop Rewriting, And Condition Splitter, Comparison Swapper, While Loop Unrolling, Double Negation Injector, Commutative Operand Permuter, Local Variable Renamer, If True Wrapper, Try Except Injector, Dead Code Injector, Dead String Assignment, Dead Method Injection, and String Literal Splitter (summarized in Table~\ref{tab:spt_catalog}, Appendix~\ref{app:spt_details}). Some mimic refactorings a developer might perform (e.g., swapping if and else branches); others stress-test the agent with unnatural or inert code (e.g., splitting string literals, injecting dead code). The two keyword-bound SPTs plant \emph{decoys}: Dead String Assignment injects an unread assignment of the keyword string, and Dead Method Injection appends an unreachable method with the keyword as its name, so an agent that searches the repository for a term from the issue also finds dead code. Appendix~\ref{app:spt_details} gives implementation details.

\paragraph{From function-level to repository-level SPTs.}
These transformation \emph{types} follow prior work on code-model robustness~\cite{ramakrishnan2022semantic, wang2023recode}, which applies them to short, self-contained functions. Each rewrite is local, but whether it preserves semantics depends on code that may lie anywhere in the repository. In a self-contained function this context is visible; in a repository it is spread across modules, and making rewrites safe there is part of our contribution. Examples of such non-local dependence include:
\begin{itemize}
    \item \textit{Name resolution.} A rewrite that introduces a call to a builtin can reach a user definition of the same name. For Loop Rewriting, for instance, introduces calls to \texttt{iter} and \texttt{next}, which a module may shadow.
    \item \textit{Implicit types.} Without type information, \texttt{a + b} may be numeric addition, which commutes, or concatenation or a user-defined \texttt{\_\_add\_\_}, which need not. Permuting operands safely requires type inference.
    \item \textit{Reflection.} Programs can inspect their own structure via \texttt{locals()}, frame introspection, metaclasses, or framework base classes (e.g., Pydantic). Even code that never runs, such as an unread assignment or a dead method, can change behavior.
\end{itemize}
Such cases are rare, but a single variant rewrites thousands of sites, so a rule that is unsafe only in rare cases is likely to meet one. Appendix~\ref{app:spt_details} lists the conditions each SPT checks.

\paragraph{Validation.}
\label{subsec:spt_validation}
We check semantics preservation empirically, in two ways. First, by differential testing~\cite{mckeeman1998differential}: we apply each transformation in isolation at every applicable site in three SWE-bench projects~\cite{jimenez2024swe} (SymPy, sqlfluff, xarray; about 43{,}000 tests), and every test kept its outcome under all 14 transformations. Second, before running an agent on a variant, we run the instance's \texttt{FAIL\_TO\_PASS} and \texttt{PASS\_TO\_PASS} tests on it and keep the variant only if they fail and pass as on the seed, as Proposition~\ref{prop:valid} requires. Both checks are limited by test coverage, so they are evidence rather than proof (Appendix~\ref{app:spt_validation}).

\section{Experiments}
\label{sec:experiments}

\subsection{Research Questions}
\label{subsec:rqs} 
Our evaluation is organized around the following questions. All three concern the untargeted generator $\Gnat$ of Section~\ref{sec:sampler}.

\begin{description}
    \item[\textbf{RQ1}]  To what extent do SPTs degrade agent performance, and how does this effect vary across models and scaffolds?
    \item[\textbf{RQ2}] How much additional effort do SPTs induce on agents, controlling for task outcome?
    \item[\textbf{RQ3}] What failure patterns emerge when agents fail on semantics-preserving variants?
\end{description}

\subsection{Setup}
\paragraph{Benchmarks and instance selection.}
We draw task instances from two 
repository-level program-repair benchmarks: SWE-bench Verified~\cite{openai2024swebenchverified}, a human-validated 500-instance subset of SWE-bench~\cite{jimenez2024swe}, and SWE-bench Pro~\cite{deng2025swe}, which contains 731 instances. Constrained by a compute budget, we 
selected 28 task instances from SWE-bench Verified and 26 from SWE-bench Pro. Our selection procedure is detailed in Appendix~\ref{sec:appendix_exp}. Each instance comes with the \texttt{FAIL\_TO\_PASS} and \texttt{PASS\_TO\_PASS} tests described in Section~\ref{subsec:lang-prims}.

\paragraph{Agents and models.}
We evaluate two 
agentic scaffolds, mini-SWE agent~\cite{yang2024sweagent} and OpenCode~\cite{opencode2024}, each driven by one of four models: Claude Opus~4.5~\cite{anthropic2025opus45}, Kimi~K2.5~\cite{team2026kimi},
MiniMax~M2.5~\cite{minimax2026m25}, and Qwen~3.6-27B~\cite{qwen36_27b}. Crossing two scaffolds, four models, and two benchmarks yields 16 configurations in total.

\subsection{Experimental Protocol}
\label{subsec:protocol}
For each instance we run the agent $20$ times on the seed and once on each of $20$ variants drawn independently from $D_i(\Gnat)$, keeping only variants that pass the validation check of Section~\ref{subsec:spt_validation}; these unperturbed and perturbed runs form the instance's two \emph{conditions}. In total, the study comprises 17{,}280 agent runs: 40 for each of the 54 instances under each of the 8 scaffold--model pairs. On our instances, the transformations in a variant target a median of 6.9\% of source lines on SWE-bench Verified and 7.7\% on SWE-bench Pro, a lower bound since each transformation also inserts new code. Every run executes in a fresh, isolated environment (Algorithm~\ref{alg:exp} in Appendix~\ref{sec:appendix_exp}). For a fixed budget of perturbed runs, spending each on a fresh variant gives the lowest-variance unbiased estimate of $\mu_i(\Gnat)$ (Appendix~\ref{sec:appendix_estimator}).

\subsection{Metrics}
\label{subsec:metrics}
All metrics are computed per configuration, a (scaffold, model, benchmark) triple; a scaffold with its model is an agent $\alpha$ (Section~\ref{subsec:lang-robustness}). A run \emph{resolves} an instance if its patch passes both the \texttt{FAIL\_TO\_PASS} and \texttt{PASS\_TO\_PASS} tests. For instance $i$, the fractions $\hat\mu_i(\Gid)$ and $\hat\mu_i(\Gnat)$ of its unperturbed and perturbed runs that resolve it estimate $\mu_i(\Gid)$ and $\mu_i(\Gnat)$. Our primary metric is the \emph{degradation} $\hat\Delta_i(\Gnat) = \hat\mu_i(\Gid) - \hat\mu_i(\Gnat)$, an estimate of $\Delta_i(\Gnat)$, which we summarize per configuration by its mean $\bar{\Delta}$ over the configuration's instances $\mathcal{I}$ (the 28 SWE-bench Verified or 26 SWE-bench Pro instances). To measure the overhead perturbation imposes on the agent, we also report the relative change in mean step count, $\delta_{\text{step}}(i) = (\bar{s}_p(i) - \bar{s}_0(i))/\bar{s}_0(i)$, where $\bar{s}_0(i)$ and $\bar{s}_p(i)$ are the mean numbers of steps (model turns or tool invocations) over the unperturbed and perturbed runs of $i$, and likewise $\delta_{\text{cost}}(i)$ for token cost, with means $\bar{\delta}_{\text{step}}$ and $\bar{\delta}_{\text{cost}}$ over $\mathcal{I}$. When conditioning on resolution, we exclude instances with no resolved baseline runs.

\paragraph{Uncertainty.}
Our inference is \emph{fixed-population}: intervals quantify run-to-run variability on the fixed instance set $\mathcal{I}$, not sampling of instances from the benchmark. For configuration-level means (degradation, step count, and cost) we report $95\%$ bootstrap percentile intervals; each resample redraws every instance's runs within each condition and keeps $\mathcal{I}$ fixed (Appendix~\ref{subsec:bootstrap}). For a single instance, degradation is a difference of two proportions estimated from 20 runs each, for which we report $95\%$ Newcombe intervals~\cite{newcombe1998interval,fagerland2015recommended}. These intervals are approximate rather than exact guarantees of the kind discussed in Section~\ref{subsec:lang-robustness}, which are conservative at our sample sizes (Appendix~\ref{subsec:est_guarantees}): bootstrap percentile intervals reach their nominal coverage only as the number of runs grows, and Newcombe intervals rely on a normal approximation that, in simulation studies, keeps coverage close to nominal even with few runs, including when all runs succeed or all fail~\cite{fagerland2015recommended}. In exchange, they are tight enough to detect the degradations we observe.

\subsection{RQ1: Degradation of Agent Performance}
\label{subsec:results}

\begin{figure}[t]
    \centering
    \includegraphics[width=\columnwidth]{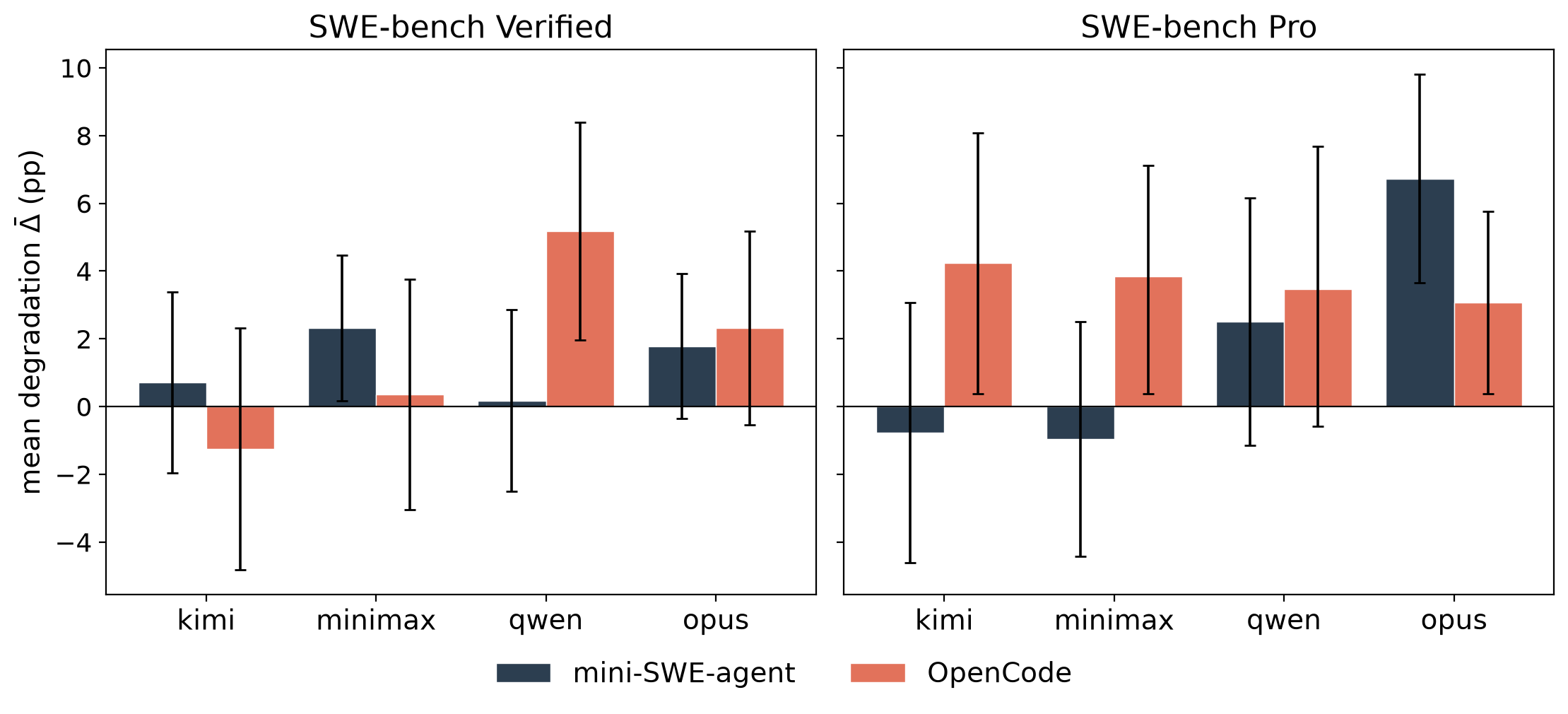}
    \caption{Mean degradation $\bar{\Delta}$ (percentage points) per model, scaffold, and benchmark. Error bars are fixed-population 95\% bootstrap confidence intervals.}
    \label{fig:mean_degradation}
\end{figure}

\paragraph{Robustness rankings do not transfer across scaffolds or benchmarks.}
Figure~\ref{fig:mean_degradation} gives $\bar{\Delta}$ for each configuration with its 95\% bootstrap interval. Perturbation reduces resolve rates in most configurations: $\bar{\Delta}$ is positive in 13 of the 16, and the interval excludes zero in 6 of the 16. No scaffold--model pair is significant on both benchmarks, so the effect is configuration-dependent rather than uniform.
No single ordering of models by robustness holds across the two scaffolds or the two benchmarks. On SWE-bench Verified, Qwen is the most robust model under mini-SWE agent at 0.2 points, yet the most brittle under OpenCode at 5.2 points. The benchmark axis is just as jagged: Opus under mini-SWE agent degrades 1.8 points on SWE-bench Verified but 6.7 points on Pro, the largest drop in the study, while under OpenCode on Pro it is the least degraded of the four models. Every model is among the most robust in at least one cell and among the most brittle in another, so $\bar{\Delta}$ measured on one scaffold and benchmark does not predict behavior on another: a practitioner who picks a model for its robustness under mini-SWE agent on SWE-bench Verified may see the opposite after switching scaffold or codebase. We call this a jagged robustness frontier: robustness is a joint property of the model, the scaffold, and the workload, not of the model alone.

\paragraph{The simpler scaffold is consistently more robust.}
Averaged over the four models, mini-SWE agent degrades less than OpenCode on both benchmarks: 1.25 points against 1.65 on SWE-bench Verified, and 1.88 against 3.65 on SWE-bench Pro. Its absolute strength, by contrast, is benchmark-dependent. It resolves 89.05\% of unperturbed runs on Verified against OpenCode's 81.20\%, but 76.35\% on Pro against OpenCode's 79.33\%. The robustness gap survives this capability flip.

\begin{figure*}[t]
    \centering
    \begin{minipage}[t]{0.48\textwidth}
        \centering
        \includegraphics[width=\linewidth]{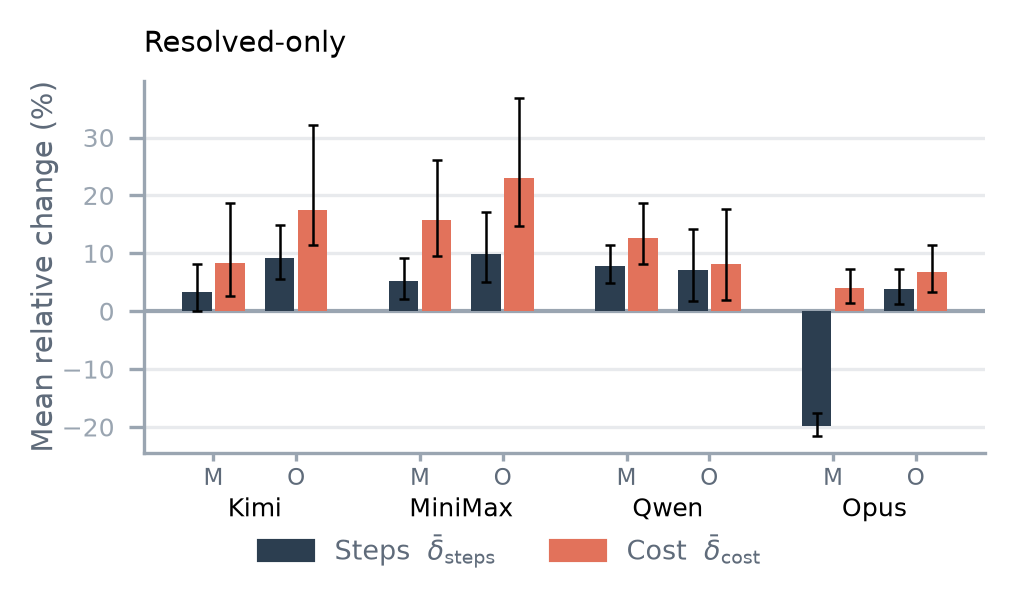}
    \end{minipage}\hfill
    \begin{minipage}[t]{0.48\textwidth}
        \centering
        \includegraphics[width=\linewidth]{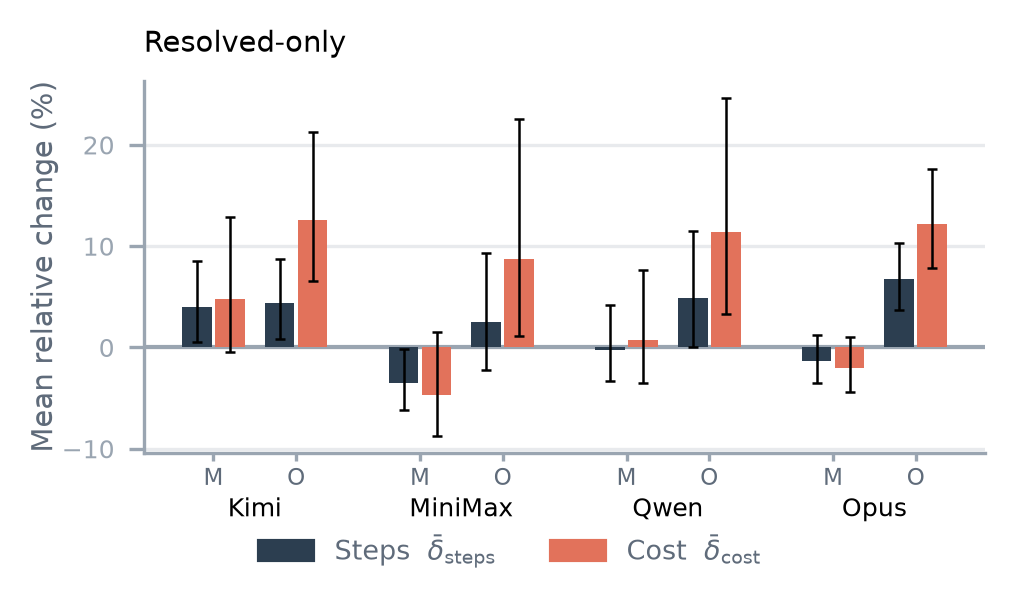}
    \end{minipage}
    \caption{Mean relative change in agent steps ($\bar{\delta}_{\text{step}}$) and cost ($\bar{\delta}_{\text{cost}}$) after perturbation, for each scaffold--model configuration on the 28 SWE-bench Verified instances (left) and the 26 SWE-bench Pro instances (right), restricted to runs the agent resolved in both cases. Positive values mean perturbed runs take more steps, or cost more, than unperturbed runs on the same instance. Error bars are fixed-population 95\% bootstrap confidence intervals.}
    \label{fig:cost_step_overhead}
\end{figure*}

\paragraph{Degradation is concentrated, not diffuse.}
Appendix~\ref{app:per_instance_degradation} reports $\hat\Delta_i(\Gnat)$ with its 95\% Newcombe interval for every instance--configuration pair. Individual comparisons are underpowered: with 20 runs per condition, only swings of roughly 25 points can exclude zero, and only 14 of the 432 pairs do. The point estimates are shifted toward positive values, and the aggregate effect is carried by a small heavy tail: three instances account for roughly two-thirds of the 6.7-point mean degradation of mini-SWE agent with Opus on SWE-bench Pro. This mirrors the jaggedness at the aggregate level. No instance degrades under all eight configurations on either benchmark, and a few cells swing sharply the other way. Brittleness attaches to particular instance--configuration pairs rather than to instances, models, or repositories alone.

\subsection{RQ2: Effort Induced by Perturbation}
\paragraph{Perturbation inflates effort more broadly than it degrades outcomes.}
Figure~\ref{fig:cost_step_overhead} reports $\bar{\delta}_{\text{step}}$ and $\bar{\delta}_{\text{cost}}$ for each configuration, restricted to the runs the agent resolved, so that the comparison is not driven by long failing runs. On SWE-bench Verified, cost rises in all 8 configurations, with intervals excluding zero. On SWE-bench Pro the increases are smaller and concentrated in the richer scaffold: all four OpenCode configurations pay more, with intervals excluding zero, while the four mini-SWE agent configurations sit near zero. The same holds over all runs (Appendix~\ref{app:effort_all_runs}): cost rises in all eight Verified configurations and seven of eight on Pro. Perturbation thus affects most configurations modestly in resolve rate but more strongly in what the agent spends.

\paragraph{The agent spends more per step, not only more steps.}
Steps and cost do not move together: cost per step rises in 12 of the 16 configurations, and on SWE-bench Verified step overhead never exceeds 9.9\% while cost overhead reaches 23.0\%. Among resolved runs, input tokens per step rise in 13 of the 16 configurations, so the agent consumes more context on each turn, consistent with longer files and heavier reading under perturbation. The scaffold contrast from RQ1 reappears on Pro: on resolved instances, the four OpenCode configurations pay 4.6--7.2\% more per step, while mini-SWE agent configurations are flat or slightly cheaper.

\paragraph{Opus compresses its trajectory rather than extending it.}
One configuration runs against this trend: Opus under mini-SWE agent on SWE-bench Verified takes \emph{fewer} steps on every perturbed instance, but emits denser actions, so its cost still rises (Appendix~\ref{app:opus_trajectory}).

\subsection{RQ3: Observed Failure Patterns}
We manually analyze 20 agent trajectories (14 distinct instances) spanning both scaffolds, both benchmarks, and all four models, to see how SPTs influence agent behavior; Appendix~\ref{subsec:appendix_selection_procedure} describes how we selected them. We describe four patterns below; Appendix~\ref{subsec:appendix_more_qual_insights} reports further observations.

\paragraph{Detecting and reverting SPTs.}
When the files the agent touches overlap substantially with the perturbed files, the agent sometimes recognizes that the code is ``obfuscated''. For example, MiniMax with mini-SWE agent on Qutebrowser checks the commit history, sees that the transformations were added in the latest commit, and reverts the repository state before making its final changes, so that all SPTs are cleared.

\paragraph{Degradation of code localization capabilities.}
SPTs can make relevant code harder to localize, either by diluting search results or by making core files look less credible. For example, Opus with mini-SWE agent on Openlibrary greps for a keyword from the issue description and is thrown off by several dead string assignments that contain the same keyword.

\paragraph{Editing or ``fixing'' perturbed code.}
Agents occasionally simplify perturbed code while patching; e.g., Opus with mini-SWE agent on Qutebrowser reports that it recognized and fixed perturbed code in several places. This extra editing enlarges the patch, raising the risk that it fails.

\paragraph{Corrupted patch validation.}
With Kimi on Pytest-dev, a unit test fails on code whose log message contains a split string literal. The agent wrongly attributes the failure to the split string, concludes its own edit is not the cause, and submits the incorrect patch without attempting a fix.

\section{Related Work}
\label{sec:related_work}

\paragraph{Robustness of code language models.}
Neural models of code are brittle to semantics-preserving perturbations, including identifier renaming~\cite{yefet2020adversarial, bielik2020adversarial, yang2022natural}, dead-code insertion and operand changes~\cite{srikant2021generating, ramakrishnan2022semantic}, and changes to docstrings, signatures, and surrounding context~\cite{wang2023recode, jha2023codeattack, mastropaolo2023robustness}; training- and architecture-based defenses improve robustness~\cite{jain2021contracode, chakraborty2022natgen, zhang2024codefort, pei2024exploiting}. Such mutations can also degrade LLMs' accuracy in predicting program outputs~\cite{orvalho2026largelanguagemodelsrobust, spiess2026robustly}. These studies evaluate single-turn inference on short inputs. We study multi-turn agents that navigate whole repositories, localize faults, and write patches, with perturbations spread across the codebase. Appendix~\ref{app:related_llm} discusses the robustness of general LLMs.

\paragraph{Specifying input distributions.}
Claims from sampling-based evaluation hold relative to the sampled distribution, which several lines of work make explicit. Statistical model checking estimates the probability that a stochastic system satisfies a property from sampled executions~\cite{younes2002probabilistic, legay2010statistical}. Statistical certificates for LLMs likewise hold over specified distributions~\cite{chaudhary2025certifying, wang2026catastrophic, yeon2025quantifying}, and Lumos~\cite{chaudhary2025lumos} provides a probabilistic programming language for writing such distributions over prompts. Property-based testing pairs programmable generators of random inputs with executable properties~\cite{claessen2000quickcheck, lampropoulos2017beginner}, and metamorphic testing~\cite{chen1998metamorphic, chen2018metamorphic} checks relations between the outputs on related inputs, the form of our robustness property. Like Lumos and property-based generators, VDL makes the input distribution programmable, but over repositories, and guarantees that every variant poses the same task as its seed.

\section{Conclusion}
\label{sec:conclusion}
We presented VDL, a probabilistic programming language for specifying distributions over semantically equivalent variants of a repository, a library of SPTs, and an untargeted generator written in VDL, and used them for, to our knowledge, the first systematic robustness evaluation of repository-level code agents, revealing a jagged robustness frontier across models, scaffolds, and repositories. Future work includes evaluating other VDL distributions and feedback-guided search.

\section*{Limitations}
\label{sec:limitations}

\paragraph{Scale and selection.} Our evaluation covers four models and two scaffolds on 54 instances sampled from two benchmarks. We selected instances that Claude Opus 4.5 resolves (Appendix~\ref{sec:appendix_exp}) so that degradation is measurable; this may raise Claude's baseline, but our findings concern changes in resolve rate under perturbation, not absolute rates. 

\paragraph{Transformation coverage.}Our SPTs cover only Python and we evaluate them under one generator setting (Section~\ref{sec:sampler}). SPTs for other languages and different generator settings may produce variants of different difficulty; VDL makes such settings straightforward to specify (Appendix~\ref{app:alt_generators}). 

\paragraph{Model-specific keyword extraction.} Keywords are extracted by the model under evaluation, so that decoys reflect what each model finds salient in the issue (Appendix~\ref{app:sampler_details}). The variants therefore differ slightly across models.

\paragraph{Statistical precision.} We run each agent 20 times per instance and condition, enough for aggregate estimates but limiting per-instance precision, so we base our claims on configuration-level means and report Newcombe intervals per instance.

\section*{Ethical Considerations}
\label{sec:ethics}

\paragraph{Potential misuse.}
Our transformations change how code looks without changing its functionality. So they could in principle hide malicious code. We consider this risk very low as the SPTs are simple and well known. Similar obfuscators are already publicly available. Our generator applies them at random rather than searching for the most damaging variants.

\paragraph{Data and execution.}
We use the public SWE-bench Verified and SWE-bench Pro benchmarks for research evaluation only. They contain no personal data, and no human subjects were involved. Each agent run used a fresh, isolated Docker container.

\paragraph{Licensing.}
We use SWE-bench Verified and SWE-bench Pro, the evaluation code and tooling for both of which is released under the MIT license. We use both benchmarks solely for non-commercial research evaluation, consistent with their terms. The libraries we build on are also under their respective open-source licenses: LibCST (MIT), Astroid (LGPL-2.1), NLTK (Apache-2.0), and vLLM (Apache-2.0). The mini-SWE agent and OpenCode scaffolds are also under open source licenses. Additionally, our released code is distributed under the MIT license, and benchmark data is not re-released in our repository.

\section*{Generative AI Usage Statement}
\label{sec:genai_usage}
Generative AI tools were used in the preparation of this manuscript to improve the clarity of the writing, to assist in writing Python scripts for the experiments, and to develop the VDL language and the arguments in Appendix F. Everything was carefully reviewed by the authors and they remain fully accountable for this work.

\bibliography{references}
\clearpage

\appendix
\section{Formal Definition of VDL}
\label{app:lang}

This appendix gives the types and typing rules of the basic VDL constructs (Section~\ref{subsec:lang-syntax}), its formal semantics, and the proof of Proposition~\ref{prop:valid}. The syntactic sugar of Figure~\ref{fig:sugar} is defined in terms of these constructs, so it needs no separate rules. Throughout, the instance $i$ is fixed, and with it $\Cseed$, $\Fgold$, and $\Kmap$.

\paragraph{Types.}
Values have the types
\[
\tau ::= \srt{Tr} \mid \srt{Kw} \mid \srt{File} \mid \srt{Site} \mid \srt{Num} \mid \srt{Bool} \mid \srt{Set}(\tau),
\]
denoting transformations in $\mathcal{T}$, keywords together with $\bot$, files, candidate sites, nonnegative real numbers and $\infty$, with $0 \cdot \infty = 0$, Booleans, and finite sets. A typing context $\Gamma$ assigns one type to each variable of the program. Expressions are typed in the standard way; in particular, $\mathcal{T}$ and $\Tkw$ have type $\srt{Set}(\srt{Tr})$, $\Src(\Cseed)$ and $\Fgold$ have type $\srt{Set}(\srt{File})$, $\Kmap(e)$ has type $\srt{Set}(\srt{Kw})$ when $e : \srt{Tr}$, and $\cand(e_1, e_2, e_3)$ has type $\srt{Set}(\srt{Site})$ when $e_1 : \srt{Tr}$, $e_2 : \srt{File}$, and $e_3 : \srt{Kw}$.

\paragraph{Typing rules.}
Figure~\ref{fig:typing} gives the rules for distributions and generators. $\Gamma \vdash d : \tau$ states that $d$ draws values of type $\tau$, and $\Gamma \vdash g$ states that $g$ is well typed. A generator is well formed if $\Gamma \vdash g$ for some $\Gamma$.

\begin{figure*}[t]
\small
\begin{mathpar}
\inferrule{\Gamma \vdash e_1 : \srt{Num} \\ \Gamma \vdash e_2 : \srt{Set}(\tau)}{\Gamma \vdash \kw{choose}(e_1, e_2) : \srt{Set}(\tau)}
\and
\inferrule{\Gamma \vdash e_1 : \srt{Num} \\ \Gamma \vdash e_2 : \srt{Num}}{\Gamma \vdash \kw{unif}(e_1, e_2) : \srt{Num}}
\and
\inferrule{ }{\Gamma \vdash \kw{skip}}
\and
\inferrule{\Gamma \vdash g_1 \\ \Gamma \vdash g_2}{\Gamma \vdash g_1 \mathbin{;} g_2}
\\
\inferrule{\Gamma \vdash e : \Gamma(x)}{\Gamma \vdash x := e}
\and
\inferrule{\Gamma \vdash d : \Gamma(x)}{\Gamma \vdash x \sim d}
\and
\inferrule{\Gamma \vdash e : \srt{Bool} \\ \Gamma \vdash g_1 \\ \Gamma \vdash g_2}{\Gamma \vdash \kw{if}\ e\ \kw{then}\ g_1\ \kw{else}\ g_2}
\and
\inferrule{\Gamma \vdash e : \srt{Set}(\Gamma(x)) \\ \Gamma \vdash g}{\Gamma \vdash \kw{for}\ x \in e\ \kw{do}\ g}
\and
\inferrule{\Gamma \vdash e_1 : \srt{Tr} \\ \Gamma \vdash e_2 : \srt{File} \\ \Gamma \vdash e_3 : \srt{Set}(\srt{Site}) \\ \Gamma \vdash e_4 : \srt{Kw}}{\Gamma \vdash \rw(e_1, e_2, e_3, e_4)}
\end{mathpar}
\caption{Typing rules for the basic VDL constructs.}
\label{fig:typing}
\end{figure*}

\paragraph{Semantics.}
A state is a pair $(C, \sigma)$ of a repository and a store $\sigma$ mapping each variable to a value of its type; every variable initially holds a default value of its type. An expression $e$ denotes a value $\denote{e}(C, \sigma)$; it depends on $C$ only through $\cand(t, f, k)$, which is interpreted as $\cand(C, t, f, k)$. A distribution $d$ denotes a probability distribution $\denote{d}(C, \sigma)$: $\kw{choose}(e_1, e_2)$ assigns probability $1/\binom{|X|}{m}$ to each subset of $X = \denote{e_2}(C,\sigma)$ of size $m = \min(\lfloor \denote{e_1}(C,\sigma) \rfloor, |X|)$, and $\kw{unif}(e_1, e_2)$ is uniform on $[\denote{e_1}(C,\sigma), \denote{e_2}(C,\sigma)]$ (the point mass on $\denote{e_1}(C,\sigma)$ unless $\denote{e_1}(C,\sigma) < \denote{e_2}(C,\sigma) < \infty$). A generator denotes a function $\denote{g}$ from states to probability distributions over states, defined in Figure~\ref{fig:semantics}. For a distribution $\mu$ and a function $\kappa$ from values to distributions, $\mu \bind \kappa$ is the distribution of $y$ obtained by drawing $v \sim \mu$ and then $y \sim \kappa(v)$; $\kappa_1 \mathbin{\fatsemi} \kappa_2$ is $s \mapsto \kappa_1(s) \bind \kappa_2$; and $\delta_s$ is the point mass on $s$. The variant distribution $D_i(g)$ is the distribution of the repository component of $\denote{g}(\Cseed, \sigma_0)$, where $\sigma_0$ is the initial store.

\begin{figure*}[t]
\small
\begin{align*}
\denote{\kw{skip}}(C, \sigma) &= \delta_{(C, \sigma)} &
\denote{g_1 \mathbin{;} g_2} &= \denote{g_1} \mathbin{\fatsemi} \denote{g_2} \\
\denote{x := e}(C, \sigma) &= \delta_{(C, \sigma[x \mapsto \denote{e}(C, \sigma)])} &
\denote{x \sim d}(C, \sigma) &= \denote{d}(C, \sigma) \bind \big(v \mapsto \delta_{(C, \sigma[x \mapsto v])}\big)
\end{align*}
\begin{gather*}
\denote{\kw{if}\ e\ \kw{then}\ g_1\ \kw{else}\ g_2}(C, \sigma) = \begin{cases} \denote{g_1}(C, \sigma) & \text{if } \denote{e}(C, \sigma) \\ \denote{g_2}(C, \sigma) & \text{otherwise} \end{cases} \\
\denote{\kw{for}\ x \in e\ \kw{do}\ g}(C, \sigma) = \big(\denote{x := v_1 \mathbin{;} g} \mathbin{\fatsemi} \cdots \mathbin{\fatsemi} \denote{x := v_m \mathbin{;} g}\big)(C, \sigma) \\
\denote{\rw(e_1, e_2, e_3, e_4)}(C, \sigma) = \begin{cases} \delta_{(\rw(C, t, f, S \cap \cand(C, t, f, k), k),\ \sigma)} & \text{if } f \in \Src(C) \\ \delta_{(C, \sigma)} & \text{otherwise} \end{cases}
\end{gather*}
\caption{Semantics of the basic VDL constructs. In the rule for $\kw{for}$, $v_1, \dots, v_m$ enumerate $\denote{e}(C, \sigma)$ in the fixed order, the set is evaluated once before the loop starts, and the empty composition is $\denote{\kw{skip}}$. In the rule for $\rw$, $t$, $f$, $S$, and $k$ are the values of $e_1, \dots, e_4$ in $(C, \sigma)$.}
\label{fig:semantics}
\end{figure*}

\noindent Every generator terminates: sets are finite, a loop's set is evaluated once before the loop starts, and VDL has no other form of repetition. For the generators in this paper, random choices affect the repository only through finitely many outcomes, since a value drawn by $\kw{unif}$ is used only in comparisons, so $D_i(g)$ is a probability distribution over finitely many repositories. Because plain transformations ignore their keyword, $\Kmap(t)$ is empty for them, and $\kw{rewrite}$ intersects $S$ with the candidate sites and does nothing outside $\Src(C)$, every well-typed generator has a meaning.

\paragraph{Proof of Proposition~\ref{prop:valid}.}
We show, by induction on the structure of $g$, that for every state $(C, \sigma)$ with $C$ having the same files as $\Cseed$, $\denote{g}(C, \sigma)$ assigns probability 1 to the set of states $(C', \sigma')$ that satisfy (a)~$C'$ has the same files and test files as $C$, and (b)~$C' \equiv C$. For $\kw{skip}$, assignments, and draws, $C' = C$. If $\mu$ assigns probability 1 to a set $A$ of states, and $\kappa(v)$ assigns probability 1 to a set $B$ for every $v \in A$, then $\mu \bind \kappa$ assigns probability 1 to $B$; together with the transitivity of $\equiv$, this gives the cases for sequencing, conditionals, and loops, which are finite compositions. For $\rw$, either $C' = C$, or $C' = \rw(C, t, f, S', k)$ with $f \in \Src(C)$ and $S' = S \cap \cand(C, t, f, k) \subseteq \cand(C, t, f, k)$. A rewrite changes only the contents of the source file $f$, which gives (a), and $C' \equiv C$ because $t$ is semantics-preserving (Definition~\ref{def:spt}), which gives (b). Taking $C = \Cseed$, a repository produced with positive probability must satisfy (a) and (b), which proves the proposition.

\section{Untargeted Generator Details}
\label{app:sampler_details}
Algorithm~\ref{alg:sampler} gives the untargeted generator of Section~\ref{sec:sampler} as pseudocode, and Algorithm~\ref{alg:assign} gives its subroutine \Call{AssignTargetNames}{}. Algorithm~\ref{alg:sampler} draws all $N$ variants of an instance; the generator $\Gnat$ describes how one of them is drawn.

\begin{algorithm*}[t]
\caption{Untargeted Generator}
\label{alg:sampler}
\small
\begin{algorithmic}[1]
\Require Seed repository $C_{\text{seed}}$, Sample count $N$, List of transformations $\mathcal{T}$, Files modified by the gold patch $\mathcal{F}_{\text{gold}}$, Issue description $\mathit{desc}$, Number of transformations to apply $N_t$, Maximum number of keywords selected $N_k$, Fraction of candidates transformed $\phi$, Maximum files per keyword-bound transformation $N_f$
\Ensure Variant population $\mathcal{V}$

\Statex
\Procedure{GenerateVariants}{$C_{\text{seed}}$, $N$, $\mathcal{T}$, $\mathcal{F}_{\text{gold}}$, $\mathit{desc}$, $N_t$, $N_k$, $\phi$, $N_f$}
    \State $\mathcal{V} \leftarrow \emptyset$
    \State $\mathcal{K} \leftarrow \textsc{None}$ 
    \For{$j \leftarrow 1$ \textbf{to} $N$}
        \State $C_{mut} \leftarrow \Call{Clone}{C_{\text{seed}}}$
        \State $T_{\text{sel}} \leftarrow \Call{RandomSelect}{\mathcal{T}, N_t}$
        \State $p(t) \sim \kw{unif}(0,1)$ \textbf{for each} $t \in T_{\text{sel}}$
        \If{$T_{\text{sel}} \cap \mathcal{T}_{\text{kw}} \neq \emptyset$ \textbf{and} $\mathcal{K} = \textsc{None}$}
            \State $\mathcal{K} \leftarrow \Call{ExtractKeywords}{\mathit{desc}}$ \Comment{LLM call; cached}
        \EndIf
        \State $\mathcal{B} \leftarrow \Call{AssignTargetNames}{T_{\text{sel}}, \mathcal{K}, N_k}$
        \For{\textbf{each} $(t, k) \in \mathcal{B}$}
            \State $c \leftarrow 0$ 
            \For{\textbf{each} $f \in \Call{SourceFiles}{C_{\text{mut}}}$}
                \State \textbf{if} $f \notin \mathcal{F}_{\text{gold}}$ \textbf{and} $\big(\Call{Random}{} > p(t)$ \textbf{or} $(t \in \mathcal{T}_{\text{kw}}$ \textbf{and} $c \geq N_f)\big)$ \textbf{then continue}
                \State $c \leftarrow c + 1$
                \State $S \leftarrow \Call{GetCandidates}{f, t, k}$
                \State \textbf{if} $S = \emptyset$ \textbf{then continue}
                \State $S' \leftarrow \Call{RandomSelect}{S, \lceil \phi |S| \rceil}$
                \State $C_{mut}[f] \leftarrow \Call{ApplyTransform}{f, S', t, k}$
            \EndFor

        \EndFor
        \State $\mathcal{V}.\text{append}(C_{mut})$
    \EndFor
    \State \Return $\mathcal{V}$
\EndProcedure
\end{algorithmic}
\end{algorithm*}

\paragraph{Notation.} $\Call{Random}{} \sim \kw{unif}(0,1)$ is a uniform random number generator. $\Call{RandomSelect}{X, m}$ returns $m$ elements drawn uniformly without replacement from $X$. $\Call{SourceFiles}{C}$ returns the source files of $C$, excluding test suite files. $\Call{Clone}{C}$ copies repository $C$. $\Call{ExtractKeywords}{\mathit{desc}}$ invokes an LLM to obtain the keyword map, $\mathcal{K}$, where $\mathcal{K}$ holds keyword candidates for $t \in \mathcal{T}_{\text{kw}}$. $\Call{AssignTargetNames}{\cdot}$ binds each selected $t \in \mathcal{T}_{\text{kw}}$ to up to $N_k$ of its candidates, emitting one pair $(t,k)$ per keyword; every other $t$, and any $t$ with no candidates, is emitted as $(t,\bot)$ (Algorithm~\ref{alg:assign}). $\Call{GetCandidates}{f, t, k}$ returns sites in $f$ where $t$ can be applied. 
$\Call{ApplyTransform}{f, S', t, k}$ rewrites $f$ at every site in $S'$.

The population $\mathcal{V}$ returned by Algorithm~\ref{alg:sampler} is consumed by the agent evaluation loop (Algorithm~\ref{alg:exp} in Appendix~\ref{sec:appendix_exp}), which provisions an isolated environment for each variant, runs the agent, and evaluates its patch against the instance's test oracle.

\paragraph{Correspondence with $\Gnat$.}
The generator follows Algorithm~\ref{alg:sampler} step by step. $\kw{choose}(N_t, \mathcal{T})$ is $\Call{RandomSelect}{\mathcal{T}, N_t}$, and $p$ is the file-inclusion probability $p(t)$, shared by every pass of a keyword-bound transformation. $\kw{targets}(t, N_k)$ is \Call{AssignTargetNames}{} (Algorithm~\ref{alg:assign}) restricted to one transformation: it draws up to $N_k$ keywords from $\Kmap(t)$, or yields $\bot$ if there are none, and plain transformations receive $\bot$. Each pair $(t, k) \in \mathcal{B}$ becomes one $\kw{scan}$ followed by one $\kw{edit}$. $\kw{scan}$ reproduces the file filter and the counter $c$ of the inner loop, with the cap $N_f$ applied only to keyword-bound transformations, and $\kw{edit}$ reproduces the selection of $\lceil \phi |S| \rceil$ candidate sites and the rewrite.

Writing the procedure as a VDL generator makes three aspects of Algorithm~\ref{alg:sampler} explicit.
\begin{itemize}
    \item \textbf{Keyword map.} \Call{ExtractKeywords}{} queries an LLM the first time a keyword-bound transformation is selected and caches the result for the instance; $\Gnat$ reads $\Kmap$ from the instance. Extraction does not depend on the generator's other random choices, so extracting once in advance gives the same distribution of variants. The $N$ variants are independent draws from $D_i(\Gnat)$ given $\Kmap$. Since the model under evaluation extracts $\Kmap$, the variant distribution varies by model.
    \item \textbf{Order.} Algorithm~\ref{alg:sampler} iterates over the set $\mathcal{B}$ without fixing an order, whereas $\kw{for}$ uses a fixed order (Section~\ref{subsec:lang-syntax}). The order matters, because one rewrite can create or remove candidate sites for another. Algorithm~\ref{alg:sampler} also interleaves the file-inclusion draws with rewriting, while $\kw{scan}$ makes them first; this leaves the distribution unchanged because the draws do not depend on file contents.
    \item \textbf{Cap accounting.} The counter $c$ is incremented for every selected file, including gold-patch files and files with no candidate site. A keyword-bound transformation bound to $m$ keywords can therefore select up to $m N_f$ files outside $\Fgold$.
\end{itemize}

Algorithm~\ref{alg:assign} describes the sub-procedure invoked by Algorithm~\ref{alg:sampler}.
 
\begin{algorithm*}[t]
\caption{Target Assignment}
\label{alg:assign}
\begin{algorithmic}[1]
\Require Selected transformations $T_{\text{sel}}$,  Keyword map $\mathcal{K}$, Maximum number of keywords to select $N_k$
\Ensure Bound (transformation, target) pairs $\mathcal{B}$
\Procedure{AssignTargetNames}{$T_{\text{sel}}, \mathcal{K}, N_k$}
\State $\mathcal{B} \leftarrow \emptyset$
\For{\textbf{each} transformation $t \in T_{\text{sel}}$}
    \If{$t \notin \mathcal{T}_{\text{kw}}$ \textbf{or} $\mathcal{K} = \textsc{None}$}
        \State $\mathcal{B} \leftarrow \mathcal{B} \cup \{ (t, \bot) \}$ \Comment{no target required}
    \Else
        \State $\mathit{names} \leftarrow \mathcal{K}(t)$ \Comment{candidates extracted for $t$}
        \If{$\mathit{names} = \emptyset$}
            \State $\mathcal{B} \leftarrow \mathcal{B} \cup \{ (t, \bot) \}$
        \Else
            \State $m \leftarrow \min(N_k, |\mathit{names}|)$
            \State $\mathit{keywords} \leftarrow \Call{RandomSelect}{\mathit{names}, m}$
            \For{\textbf{each} $k \in \mathit{keywords}$}
                \State $\mathcal{B} \leftarrow \mathcal{B} \cup \{ (t, k) \}$
            \EndFor
        \EndIf
    \EndIf
\EndFor
\State \Return $\mathcal{B}$
\EndProcedure
\end{algorithmic}
\end{algorithm*}

\section{Further Alternative Generators}
\label{app:alt_generators}

This appendix describes alternatives to $\Gnat$ (Section~\ref{sec:sampler}). All but the last change a single aspect of $\Gnat$, so that a difference between $\Delta_i(g)$ and $\Delta_i(\Gnat)$ can be attributed to that aspect; the last, $g_{\text{flood}}$, is a stress test that changes several at once. We write $\Gnat[x := v]$ for $\Gnat$ with the hyperparameter or subexpression $x$ replaced by $v$. We have not run these generators. Figure~\ref{fig:alt_generators} gives them in VDL.

\begin{figure*}[t]
\small
\begin{align*}
g_t &\triangleq p \sim \kw{unif}(0,1) \mathbin{;} \kw{if}\ t \in \Tkw\ \kw{then}\ \big(W \sim \kw{targets}(t, N_k) \mathbin{;} \kw{for}\ k \in W\ \kw{do}\ (F \sim \kw{scan}(p, N_f) \mathbin{;} \kw{edit}(t, k, F, \phi))\big) \\
&\quad \kw{else}\ \big(F \sim \kw{scan}(p, \infty) \mathbin{;} \kw{edit}(t, \bot, F, \phi)\big) \\
\noalign{\vskip 3pt\hrule\vskip 3pt}
g_{\phi'} &\triangleq T_{\text{sel}} \sim \kw{choose}(N_t, \mathcal{T}) \mathbin{;} \kw{for}\ t \in T_{\text{sel}}\ \kw{do}\ \Big( \\
&\quad p \sim \kw{unif}(0,1) \mathbin{;} \kw{if}\ t \in \Tkw\ \kw{then}\ \big(W \sim \kw{targets}(t, N_k) \mathbin{;} \kw{for}\ k \in W\ \kw{do}\ (F \sim \kw{scan}(p, N_f) \mathbin{;} \kw{edit}(t, k, F, \phi'))\big) \\
&\quad \kw{else}\ \big(F \sim \kw{scan}(p, \infty) \mathbin{;} \kw{edit}(t, \bot, F, \phi')\big)\Big) \\
\noalign{\vskip 3pt\hrule\vskip 3pt}
g_{\text{on}} &\triangleq T_{\text{sel}} \sim \kw{choose}(N_t, \mathcal{T}) \mathbin{;} \kw{for}\ t \in T_{\text{sel}}\ \kw{do}\ \Big( \\
&\quad \kw{if}\ t \in \Tkw\ \kw{then}\ \big(W \sim \kw{targets}(t, N_k) \mathbin{;} \kw{for}\ k \in W\ \kw{do}\ \kw{edit}(t, k, \Fgold, \phi)\big)\ \kw{else}\ \kw{edit}(t, \bot, \Fgold, \phi)\Big) \\
\noalign{\vskip 3pt\hrule\vskip 3pt}
g_{\text{off}} &\triangleq T_{\text{sel}} \sim \kw{choose}(N_t, \mathcal{T}) \mathbin{;} \kw{for}\ t \in T_{\text{sel}}\ \kw{do}\ \Big( \\
&\quad p \sim \kw{unif}(0,1) \mathbin{;} \kw{if}\ t \in \Tkw\ \kw{then}\ \big(W \sim \kw{targets}(t, N_k) \mathbin{;} \kw{for}\ k \in W\ \kw{do}\ (F \sim \kw{scan}(p, N_f) \mathbin{;} \kw{edit}(t, k, F \setminus \Fgold, \phi))\big) \\
&\quad \kw{else}\ \big(F \sim \kw{scan}(p, \infty) \mathbin{;} \kw{edit}(t, \bot, F \setminus \Fgold, \phi)\big)\Big) \\
\noalign{\vskip 3pt\hrule\vskip 3pt}
g_{\text{perv}} &\triangleq T_{\text{sel}} \sim \kw{choose}(N_t, \mathcal{T}) \mathbin{;} \kw{for}\ t \in T_{\text{sel}}\ \kw{do}\ \Big( \\
&\quad p := 0.9 \mathbin{;} \kw{if}\ t \in \Tkw\ \kw{then}\ \big(W \sim \kw{targets}(t, N_k) \mathbin{;} \kw{for}\ k \in W\ \kw{do}\ (F \sim \kw{scan}(p, N_f) \mathbin{;} \kw{edit}(t, k, F, \phi))\big) \\
&\quad \kw{else}\ \big(F \sim \kw{scan}(p, \infty) \mathbin{;} \kw{edit}(t, \bot, F, \phi)\big)\Big) \\
\noalign{\vskip 3pt\hrule\vskip 3pt}
g_{\text{ref}} &\triangleq T_{\text{sel}} \sim \kw{choose}(N_t, \mathcal{T}_{\text{ref}}) \mathbin{;} \kw{for}\ t \in T_{\text{sel}}\ \kw{do}\ \big(p \sim \kw{unif}(0,1) \mathbin{;} F \sim \kw{scan}(p, \infty) \mathbin{;} \kw{edit}(t, \bot, F, \phi)\big) \\
\noalign{\vskip 3pt\hrule\vskip 3pt}
g_{\text{flood}} &\triangleq \kw{for}\ t \in \{t_{\text{ds}}, t_{\text{dm}}\}\ \kw{do}\ \big(W \sim \kw{targets}(t, 20) \mathbin{;} \kw{for}\ k \in W\ \kw{do}\ \kw{edit}(t, k, \Fgold, 1)\big)
\end{align*}
\caption{The alternative generators of Appendix~\ref{app:alt_generators} in VDL, using the syntactic sugar of Figure~\ref{fig:sugar}. All but $g_{\text{flood}}$ differ from $\Gnat$ (Figure~\ref{fig:galg}) in one aspect. $g_{\text{ref}}$ needs no keyword branch because $\mathcal{T}_{\text{ref}}$ contains no keyword-bound transformation.}
\label{fig:alt_generators}
\end{figure*}

\paragraph{Per-transformation ablation.}
For each $t \in \mathcal{T}$, $g_t = \Gnat[\mathcal{T} := \{t\}, N_t := 1]$ applies only $t$. The hypothesis is that a few transformations account for most of the degradation, mirroring its concentration in a few instances (Section~\ref{subsec:results}).

\paragraph{Dose and response.}
The generators $g_{\phi'} = \Gnat[\phi := \phi']$ for $\phi' \in \{0.1, 0.3, 0.5, 0.7, 0.9\}$ vary the fraction of candidate sites rewritten while keeping the mix of transformations. The hypothesis is that resolve rates hold up to a threshold and decline beyond it; $\Gnat$ measures one point on it.

\paragraph{Locality.}
In $\Gnat$, every pass rewrites the gold-patch files together with a random share of the other files. $g_{\text{off}}$ replaces the file argument $F$ of every $\kw{edit}$ by $F \setminus \Fgold$, so no gold-patch file is touched; degradation under $g_{\text{off}}$ would come from the surrounding code rather than from difficulty reading or localizing the code to be fixed. $g_{\text{on}}$ replaces the file argument $F$ of every $\kw{edit}$ by $\Fgold$, so only gold-patch files are rewritten, and $g_{\text{perv}} = \Gnat[\kw{unif}(0,1) := \kw{unif}(0.9,0.9)]$ fixes the inclusion probability at $0.9$, so most source files are rewritten.

\paragraph{Refactorings only.}
The catalog mixes transformations that mimic refactorings a developer might perform with transformations that introduce unnatural or inert code. Let $\mathcal{T}_{\text{ref}}$ contain If Else Switcher, For Loop Rewriting, And Condition Splitter, Comparison Swapper, Commutative Operand Permuter, and Local Variable Renamer. $g_{\text{ref}} = \Gnat[\mathcal{T} := \mathcal{T}_{\text{ref}}]$ applies only these. If $\Delta_i(g_{\text{ref}})$ is positive, agents are sensitive to how ordinary code is structured, and not only to injected inert code.

\paragraph{Keyword flooding.}
Let $t_{\text{ds}}$ and $t_{\text{dm}}$ be Dead String Assignment and Dead Method Injection. The generator $g_{\text{flood}}$ places decoys at every candidate site of the gold-patch files for up to 20 keywords, without the $N_f$ cap. It tests how agents localize the fix when the files they must edit contain many misleading matches for terms in the issue, and whether they detect and revert such variants, the behavior the cap in $\Gnat$ is designed to avoid.

\section{Semantics-Preserving Transformations}
\label{app:spt_details}

The appendix details the implementation of each SPT. It describes the applicability condition of each SPT that restricts the application of the SPT to the candidate sites. Writing these conditions is where the difficulty lies. They must hold across diverse repositories, where the same construct appears in many different implementations, so a condition has to rule out every unsafe variant rather than the common ones. Writing an SPT for a function-level code snippet is comparatively easy. The snippet is short and self-contained, its bindings and callers are visible in the code being rewritten, and the only observable behavior is the value it returns.

None of this holds in a repository. Names resolve through nested scopes and imports, so a builtin such as \texttt{iter}, \texttt{next}, or \texttt{Exception} may be shadowed at the rewrite site. Types are not written down either. Addition commutes on numbers, but not on strings, lists, or a class with a custom \texttt{\_\_add\_\_}. A program can also observe its own structure. It can do so through \texttt{locals()}, frame introspection, metaclasses, attribute-dispatch hooks, and framework base classes that fix what a class namespace may contain. Code that never runs can therefore still change behavior. Scale makes this more difficult. A single variant transforms thousands of sites in a repository. So a rule that is not semantics-preserving in a rare case will meet that case.  We therefore keep every SPT conservative. When an applicability condition cannot be established from the code, we skip the site.

\begin{table}[!t]
\centering
\footnotesize
\setlength{\tabcolsep}{3pt}
\begin{tabular}{@{}lp{3.3cm}@{}}
\toprule
\textbf{Transformation} & \textbf{Summary} \\
\midrule
If Else Switcher & Swaps if/else branches and negates the condition \\
For Loop Rewriting & Rewrites a \texttt{for} loop using an explicit iterator \\
And Condition Splitter & Decomposes \texttt{if A and B} into nested \texttt{if}s \\
Comparison Swapper & Swaps operands and inverts the operator \\
While Loop Unrolling & Unrolls one iteration of a while loop \\
Double Negation Injector & Wraps a condition in \texttt{not not ($\cdot$)} \\
Commutative Operand Permuter & Reorders commutative operands \\
Local Variable Renamer & Renames safe local variables to synonyms \\
If True Wrapper & Wraps a block in a permanently true guard \\
Try Except Injector & Wraps a block in a redundant try/except \\
Dead Code Injector & Inserts an unreachable block \\
Dead String Assignment$^\dagger$ & Inserts an unread variable assignment \\
Dead Method Injection$^\dagger$ & Appends an unreachable method to a class \\
String Literal Splitter & Splits a string literal into a concatenation \\
\bottomrule
\end{tabular}
\caption{Catalog of semantics-preserving transformations (SPTs). $^\dagger$Keyword-bound ($\Tkw$).}
\label{tab:spt_catalog}
\end{table}

We group the transformations of Table~\ref{tab:spt_catalog} by the operation they perform: rewrites of existing structure and insertions of inert code.

\subsection{Rewrites}

\paragraph{If Else Switcher} negates the condition and exchanges the two branches. It is applied only to \textbf{if} statements that have a direct \texttt{else} clause. An \texttt{if} whose \texttt{else} branch is another \texttt{if} (i.e.\ an \texttt{elif}) is not switched, although an \texttt{elif} clause carrying its own \texttt{else} is itself a valid target. This SPT is also applied to inline ternary expressions, where an \texttt{else} is syntactically required. The condition, and in the ternary case both result expressions, are wrapped in parentheses before negation to maintain correct operator precedence.

\paragraph{And Condition Splitter} decomposes a compound \texttt{if} statement whose condition is a Boolean conjunction (\texttt{A and B}) into semantically equivalent nested \texttt{if} statements. When the statement has an \texttt{else} clause, that clause is duplicated onto both levels so that failure of either conjunct reaches it.

\paragraph{Comparison Swapper} exchanges the operands of an asymmetric relational comparison \texttt{(\textless{}, \textgreater{}, \textless{}=, \textgreater{}=)} and inverts the operator to preserve semantic equivalence. It is restricted to simple two-operand comparisons: chained comparisons and symmetric operators are excluded.

\paragraph{Double Negation Injector} wraps an \texttt{if} or \texttt{while} condition in \texttt{not not (Condition)}. It skips literal \texttt{True/False} conditions and conditions that are themselves a top-level negation, so that a condition such as \texttt{not x} is left alone while a compound condition that merely contains a \texttt{not} (e.g.\ \texttt{not a and b}) remains eligible.

\paragraph{While Loop Unrolling} unrolls one iteration of a \texttt{while} loop: the loop body is duplicated at the head of the new body, followed by an inner \texttt{while} carrying the original test and body, followed by an unconditional \texttt{break}. The outer test is thus evaluated as many times as in the original loop, which matters when the test has side effects. Only loops with an indented body are transformed, and loops with an \texttt{else} clause are excluded, since the injected \texttt{break} would suppress it.

\paragraph{For Loop Rewriting} rewrites a \texttt{for} loop into a semantically equivalent \texttt{while True} loop using the explicit iterator protocol. It materializes an iterator with \texttt{iter(...)} and advances it with \texttt{next(...)} inside a \texttt{try/except StopIteration: break} block. The introduced iterator and flag names carry a random suffix so they cannot collide with existing bindings. It also emulates the \texttt{for ... else} contract via a sentinel flag. Specifically, each \texttt{break} statement belonging to the rewritten loop---breaks inside nested loops are left untouched---is rewritten to \texttt{flag = False; break}, so that the flag correctly records whether the loop exited early. If the iterable is an implicit tuple, it is explicitly parenthesized before being passed to \texttt{iter()} to avoid the two-argument \texttt{iter(callable, sentinel)} misinterpretation. The transformation is skipped when any of \texttt{iter}, \texttt{next}, or \texttt{StopIteration} is shadowed by a binding in an enclosing scope. \texttt{async for} loops are excluded, as their asynchronous protocol cannot be emulated with synchronous \texttt{iter()}/\texttt{next()}.

\paragraph{Commutative Operand Permuter} swaps the operands of commutative binary operations: addition, multiplication, bitwise AND, OR, and XOR. Because commutativity holds for these operators only over specific types in Python (e.g., string and list \texttt{+} are order-dependent, and a user-defined \texttt{\_\_add\_\_}/\texttt{\_\_radd\_\_} need not commute), the transformation performs type inference using Astroid~\cite{astroid} before swapping. The swap is applied only when both operands are independently inferred to lie within the allowed type set for that operator. If inference is inconclusive or the inferred type lies outside the allowed set, the transformation is skipped. Logical \texttt{and}/\texttt{or} are never permuted.

\paragraph{Local Variable Renamer} renames local variables to semantically related synonyms. Candidates are the targets of plain assignment statements inside function bodies. Module-, class-, lambda- and comprehension-level bindings are never renamed. Candidate names are looked up in WordNet~\cite{miller1995wordnet} for a synonym. If none is suitable, or the synonym would collide with an identifier already present in the module, a random five-letter lowercase name is generated. All occurrences within the scope of the binding are updated consistently, including nested scopes that do not rebind the name. We apply a conservative exclusion list to avoid renaming anything whose identity may be externally meaningful or whose rename could not be confined to one scope: built-ins, keywords, very short names, function names, parameters, imported and decorator names, names declared \texttt{global} or \texttt{nonlocal}, loop targets, and variables used in a call position, as an argument, or as part of an attribute access or the base of a subscript. We additionally exclude variables that alias another name (\texttt{a = b}), variables assigned the result of a call, and any variable bound more than once in its scope, since control flow then determines which binding a reference observes.

\paragraph{String Literal Splitter} fragments a string literal into a \texttt{+}-joined concatenation at a randomly chosen split point, exploiting that Python evaluates the concatenation of literals to the original value. Because the literal no longer appears as one contiguous token, this frustrates localization strategies that search for the whole string. Splitting is skipped for literals inside type annotations, prefixed byte/raw/unicode strings (escape semantics), strings shorter than a safe split length of four characters, literals used as the base of an attribute access, and literals already part of an implicit concatenation, a non-\texttt{+} binary operation, or a \texttt{match} pattern. Formatted (f-)strings are never targeted.

\subsection{Inert Insertions}

These transformations add code that would never alter program behavior. Two of them---Dead String Assignment and Dead Method Injection---insert code that is inert but lexically salient, forcing an agent that relies on surface cues such as identifier or method names to distinguish genuine edit sites from decoys.

\paragraph{If True Wrapper} encloses an indented block in a guard built from a grammar of randomly generated tautologies, whose complexity is set by a depth parameter. Because the guard always evaluates to \texttt{True}, every wrapped statement still executes. Candidate blocks are those spanning between 5 and 100 lines, and blocks whose first statement is a docstring are skipped.

\paragraph{Try Except Injector} wraps an indented block in \texttt{try ... except Exception: raise}. The \texttt{except} clause unconditionally re-raises any exception it catches, so all exceptions are propagated to the same handler as in the unmodified code. We use the \texttt{except Exception: raise} pattern instead of \texttt{except: raise} to avoid catching \texttt{BaseException} subclasses such as \texttt{KeyboardInterrupt} and \texttt{SystemExit}. As with the previous transformation, candidates are blocks of 5 to 100 lines and blocks whose first statement is a docstring are skipped; injection is additionally skipped when \texttt{Exception} is shadowed by a binding at the injection site, which would otherwise make the handler itself raise \texttt{TypeError}.

\paragraph{Dead Code Injector} inserts an \texttt{if False} block containing a dummy assignment at the head of an indented block. The guard renders the block statically unreachable. When the block begins with a docstring, the injection is placed after it so that \texttt{\_\_doc\_\_} is preserved.

\paragraph{Dead String Assignment} injects an unread assignment of the form \texttt{\textless{}name\textgreater{} = "\textless{}keyword\textgreater{}"} before selected statements inside a function body. The keyword-like string serves as a distractor resembling content an agent might search for, while remaining a no-op. The \texttt{\textless{}name\textgreater{}} is drawn from a fixed pool, chosen so as not to shadow any identifier used in the enclosing function, and is never read. Injection never displaces a leading docstring and is not placed after an unconditional terminator (\texttt{return}/\texttt{raise}/\texttt{break}/\texttt{continue}) in the same block. Any function that reflects on its local namespace is excluded, as introducing a new binding would be observable: this covers \texttt{locals()}, argument-less \texttt{vars()}, \texttt{eval}/\texttt{exec} calls that do not pass explicit globals and locals, and frame introspection via \texttt{sys.\_getframe}, \texttt{inspect.currentframe}, \texttt{inspect.stack}, and related APIs.

\paragraph{Dead Method Injection} appends a structurally complete method, \texttt{def \textless{}name\textgreater{}(self, *args, **kwargs)} whose entire body is guarded by \texttt{if False}, to a class body, serving as a highly plausible edit decoy. The method is callable but its body is unreachable, so it always returns \texttt{None} without side effects. Crucially, appending a method is not unconditionally behavior-preserving, because a class may observe its own attribute namespace. The transformation is therefore skipped whenever a class controls that namespace: classes declaring a custom metaclass, classes that override attribute dispatch (\texttt{\_\_getattr\_\_}, \texttt{\_\_getattribute\_\_}, \texttt{\_\_setattr\_\_}, \texttt{\_\_delattr\_\_}), property-only namespaces, sentinel classes whose body is a bare \texttt{pass} or \texttt{...}, and classes whose base list mentions an attribute-constraining framework type (e.g., Pydantic \texttt{BaseModel}, \texttt{Enum}, \texttt{NamedTuple}, \texttt{TypedDict}, \texttt{Protocol}, \texttt{ABC}), matched by the simple name of the base expression. A class is also skipped when it already declares the target name as a method or class attribute, and dunder target names are excluded outright.

\subsection{Validation and Composition}
\label{app:spt_validation}

\paragraph{Validation procedure.}
We validate that our SPTs are semantics-preserving empirically through differential testing \cite{mckeeman1998differential} against the test suites of three projects drawn from our experimental benchmark, SWE-bench \cite{jimenez2024swe}, spanning distinct domains: SymPy (symbolic
mathematics; 12{,}994 tests), sqlfluff (SQL linting; 10{,}060 tests), and
xarray (labeled $N$-dimensional arrays; 19{,}917 tests). Each transformation is validated in isolation: for a given transformation, we apply the transformation at every applicable site, run the full test-suite, and compare the per-test outcome against the unmodified baseline. Validating one transformation at a time renders any divergence in behavior attributable to a single transformation type. Across all three projects, every test retained its outcome under all 14 transformations. This evidence is bounded by the coverage of the underlying test-suites and is therefore not a proof of equivalence. In addition to this, every sampled variant is checked against its own instance's \texttt{PASS\_TO\_PASS} and \texttt{FAIL\_TO\_PASS} tests before use (Section~\ref{subsec:spt_validation}). This validation is the evidence that our transformations are semantics-preserving in the sense of Definition~\ref{def:spt}.

\paragraph{Composition and scope.}
Individual transformations act on candidate sites within a single file, whereas evaluating an agent on a code repair task requires perturbing a whole repository. A VDL generator (Section~\ref{sec:lang}) therefore applies a finite sequence $\langle t_1,\dots,t_m\rangle$ of SPTs to the source files, excluding test suite files, of a repository. Each $t_j$ is individually semantics-preserving, and because observational equivalence is transitive, their composition is semantics-preserving as well (Proposition~\ref{prop:valid}). The perturbed repository is therefore observationally equivalent to the original.

The exclusion of test suite files (Proposition~\ref{prop:valid}) is what keeps such a perturbed repository usable as an evaluation target. The oracle that defines task success is identical before and after perturbation, and no change in an agent's score can be attributed to a moved target.

\section{Experimental Procedure Details}
\label{sec:appendix_exp}

\subsection{Instance Sampling Procedure}
To construct our task sample, we first identified instances resolved by Claude Opus 4.5 using official SWE-bench Verified leaderboard results, ensuring that our sampled tasks were solvable under standard (unperturbed) conditions. We then performed stratified sampling over repository–difficulty combinations, drawing one instance per stratum to ensure balanced coverage of the instance space.

Since official instance wise results were unavailable for SWE-bench Pro, we started by sampling an equal number of instances from each of the repos, and running Claude Opus 4.5 on the set. Then, 26 passing instances were sampled across all repos that contained at least one valid python file to which the SPTs could be applied. 

\subsection{Full Experimental Procedure}
\label{subsec:appendix_exp_procedure}

Algorithm~\ref{alg:exp} gives the full experimental procedure summarized in the Protocol paragraph of the experiments section. For each instance it proceeds in four phases: seed retrieval, baseline evaluation ($N$ runs on the unperturbed seed), variant generation via Algorithm~\ref{alg:sampler} (the untargeted generator of Section~\ref{sec:sampler}), and perturbed evaluation (one run per variant, each in a freshly provisioned environment).

\begin{algorithm*}[t]
\caption{Agent Robustness Experiment}
\label{alg:exp}
\begin{algorithmic}[1]
\Require Set of Instance IDs $\mathcal{I}$, Number of samples $N$, Configuration $\mathit{config}$, Agent $\alpha$, Evaluator $\mathcal{E}$, Transformations $\mathcal{T}$, Generator hyperparameters $N_t, N_k, \phi, N_f$
\Ensure Baseline Reports $\mathcal{R}_0$, Perturbed Reports $\mathcal{R}_p$

\Procedure{RunExperiment}{$\mathcal{I}$, $N$, $\mathit{config}$, $\alpha$, $\mathcal{E}$, $\mathcal{T}$, $N_t$, $N_k$, $\phi$, $N_f$}
    \State $\mathcal{R}_0 \leftarrow \emptyset$, $\mathcal{R}_p \leftarrow \emptyset$ \Comment{Initialize report collections}
    \For{\textbf{each} instance $i \in \mathcal{I}$}
        \State \Comment{Phase 1: Seed Retrieval}
        \State $(C_{seed}, \mathcal{F}_{\text{gold}}, desc) \leftarrow \Call{ExtractInstance}{i}$

        \State \Comment{Phase 2: Baseline Evaluation}
        \For{$k \leftarrow 1$ \textbf{to} $N$}
            \State $E_{eval} \leftarrow \Call{ProvisionEnvironment}{i}$ \Comment{unperturbed seed, no injection}
            \State $\mathcal{R} \leftarrow \Call{RunAndEvaluate}{\alpha, \mathcal{E}, E_{eval}, i, \mathit{config}}$
            \State $\mathcal{R}_0 \leftarrow \mathcal{R}_0 \cup \mathcal{R}$
        \EndFor

        \State \Comment{Phase 3: Variant Generation (Algorithm~\ref{alg:sampler})}
        \State $\mathcal{S} \leftarrow$ \Call{GenerateVariants}{$C_{seed}$, $N$, $\mathcal{T}$, $\mathcal{F}_{\text{gold}}$, $desc$, $N_t$, $N_k$, $\phi$, $N_f$}

        \State \Comment{Phase 4: Perturbed Evaluation}
        \For{\textbf{each} sample $s \in \mathcal{S}$}
            \State $E_{eval} \leftarrow \Call{ProvisionEnvironment}{i}$
            \State $\Call{InjectCode}{E_{eval}, s}$

            \State $\mathcal{R} \leftarrow \Call{RunAndEvaluate}{\alpha, \mathcal{E}, E_{eval}, i, \mathit{config}}$
            \State $\mathcal{R}_p \leftarrow \mathcal{R}_p \cup \mathcal{R}$
        \EndFor
    \EndFor
    \State \Return $(\mathcal{R}_0, \mathcal{R}_p)$
\EndProcedure

\Statex
\Procedure{RunAndEvaluate}{$\alpha, \mathcal{E}, E, i, \mathit{config}$}
    \State $Patch \leftarrow \Call{ExecuteAgent}{\alpha, E, i, \mathit{config}}$
    \State $Report \leftarrow \Call{EvaluatePatch}{\mathcal{E},i, Patch}$
    \State \Return $Report$
\EndProcedure

\end{algorithmic}
\end{algorithm*}

\subsection{Computational Infrastructure}

All experiments were run on one lab server. It has two AMD EPYC 9554 64-core processors (128 cores, 256 threads), 768 GB RAM, 6 Nvidia RTX PRO 6000 Blackwell GPUs (96 GB VRAM per GPU), and a 7 TB NVMe drive. The operating system is Ubuntu 24.04.4 LTS. Each agent run executes in its own Docker container (Docker version 29.5.2).

We serve Qwen 3.6-27B on this machine with vLLM (v0.20.1) using 2 GPUs. We access Claude Opus 4.5 (AWS Bedrock ID: anthropic.claude-opus-4-5-20251101-v1:0), Kimi K2.5 (AWS Bedrock ID: moonshotai.kimi-k2.5) and Minimax M2.5 (AWS Bedrock ID: minimax.minimax-m2.5) using Amazon Bedrock API. All 4 models used temperature 1.0. Our code is Python 3.10, using LibCST v1.8.5 for SPTs, Astroid v3.3.0 for type inference, and NLTK v3.6.0 for WordNet lookups. The scaffolds are mini-SWE agent v1.17.5 and OpenCode v1.18.4.

\paragraph{Cost accounting of Qwen 3.6.}
Reported costs are token costs. For the three models accessed through Bedrock we use the provider's published per-token prices. Qwen 3.6-27B is served locally and to keep costs comparable across models we price its input and output tokens at the OpenRouter rates for the same model. The same price schedule is applied to the unperturbed and perturbed runs of an instance, so $\delta_{\text{cost}}$ compares runs within an instance consistently. 

\subsection{Bootstrap Intervals}
\label{subsec:bootstrap}
For the configuration-level means (degradation, step count, and cost) we report $95\%$ bootstrap percentile intervals from $B = 20{,}000$ resamples, resampling runs within each instance while leaving the instance set intact. For degradation, we redraw each instance's per-run outcomes in each condition from a binomial model, $\mathrm{Binomial}(n, \hat\mu_i(g))/n$, where $g$ is $\Gid$ or $\Gnat$ and $n$ is the number of runs in that condition. For step count and cost, we draw the per-run values within each condition with replacement and recompute the instance's relative change.

\subsection{Randomness and Reproducibility}
\label{subsec:appendix_randomness}

We do not fix a random seed. Our pipeline is random in two places. The untargeted generator (Algorithm~\ref{alg:sampler}) draws the transformation subset, the file-inclusion probabilities, the sites within each file, and the target keywords. And all four models are sampled at temperature 1.0, so a run is stochastic even on a fixed repository. A seed fixes which 20 variants we draw, not what the agent does on them: rerunning the agent on the same variant can give a different outcome. Reproducing our results therefore means drawing 20 fresh variants per instance at the same hyperparameters and recovering our reported effects. The bootstrap of the statistical protocol is the exception: it is seeded.

\subsection{Code and Instance Lists}
\label{subsec:appendix_artifacts}

The code contains the full implementation of the experiment and analysis. This includes the SPT implementations, the untargeted generator of Algorithm~\ref{alg:sampler}, the evaluation harness of Algorithm~\ref{alg:exp}, the analysis scripts that produce the reported statistics and figures, and an interactive dashboard for exploring the results. The prompts used are included there as well, as is the list of the 28 SWE-bench Verified and 26 SWE-bench Pro instance IDs used in the study.

\subsection{Agent Scaffold Comparison}

Table~\ref{tab:sacffold_comparison} compares the mini-SWE agent and OpenCode scaffolds.
\begin{table*}[h]
    \centering
    \small
    \setlength{\tabcolsep}{4pt}
    \renewcommand{\arraystretch}{1.4}
    \begin{tabular}{@{} l p{4.2cm} p{8cm}@{}}
        \toprule
        \textbf{Category} & \textbf{mini-SWE} & \textbf{OpenCode} \\
        \midrule
        Architecture & Single, centralized agent & Primary agents invoke tools and subagents for specific tasks (e.g., Build, Plan, General, Explore) \\
        Available Tools & \texttt{bash} & \texttt{bash}, \texttt{edit}, \texttt{write}, \texttt{read}, \texttt{grep}, \texttt{glob}, \texttt{lsp}, \texttt{apply\_patch}, \texttt{skill}, \texttt{todo\_write}, \texttt{web\_fetch}, \texttt{web\_search}, \texttt{question} \\
        Access Restrictions & None & Available tools determined by an agent's role and user configuration \\
        Context Management & Continuous, append-only & Multiple, dynamically changing contexts for different subagents\\
        \bottomrule
    \end{tabular}
    \caption{Comparison of mini-SWE and OpenCode Agent Scaffolds}
    \label{tab:sacffold_comparison}
\end{table*}

\subsection{Selection of hyperparameter values}
\label{subsec:hyperparams}
We did not conduct a search over the generator's hyperparameter space. A single configuration at a single hyperparameter setting costs $54 \times 40 = 2{,}160$ agent runs. So even a coarse grid over $(N_t, N_k, \phi, N_f)$ would multiply the cost of the study by an order of magnitude. We instead fixed values for each parameter. $N_t = 3$ of the 14 transformations makes a variant a composition of several SPTs. $\phi = 0.7$ leaves a minority of candidate sites in a file untouched. So a variant contains both transformed and untransformed instances of the same pattern. $N_k = 5$ and $N_f = 10$ are the only values set from observation. Without them, agents often detected the perturbation and reverted the repository (e.g.\ via \texttt{git reset}), which destroys the measurement rather than making it harder.

\section{Unbiasedness of the Degradation Estimator}
\label{sec:appendix_estimator}

We estimate perturbed performance by running the agent
\emph{once} on each of $N=20$ sampled variants. Because a single run of a stochastic agent is a noisy measurement of that variant's difficulty, it is reasonable to ask
whether one run per variant is enough. An alternative would be to draw fewer variants and run each several times, averaging away the agent's randomness. This appendix shows that no such averaging is needed. Per-instance degradation is unbiased for  \emph{any} number of runs per variant including one, and under a fixed run budget one run per variant is the variance-minimizing allocation.

\subsection{Setup and notation}
\label{subsec:est_setup}

Fix a task instance $i$ and an agent $\alpha$, and use the notation of Section~\ref{subsec:lang-robustness}.

\paragraph{Variant sampling.} A variant $V$ is drawn from $D_i(\Gnat)$. Its resolve probability $p(i[V])$ is the probability that a run of $\alpha$ resolves instance $i$ with its seed repository replaced by $V$, taken over the agent's own randomness. Because $V$ is itself random, $p(i[V])$ is a random variable. It
is the difficulty of a randomly chosen semantically equivalent rewrite of instance $i$.

\paragraph{Agent sampling.} Given the variant $V$, a single agent run yields a resolve indicator $Y$ with
\begin{equation}
    \begin{gathered}
    Y \mid V \;\sim\; \mathrm{Bernoulli}(p(i[V])), \\
    \mathbb{E}[Y \mid V] \;=\; p(i[V]).
    \end{gathered}
    \label{eq:bernoulli}
\end{equation}
The agent's stochasticity is conditionally independent of how the variant was chosen, given the variant. This is enforced by the experimental protocol. Every run executes in a freshly provisioned, isolated environment, so no state carries between runs.

\paragraph{Estimands.} The quantity that $\hat\mu_i(\Gnat)$ estimates is the mean per-variant resolve probability,
\begin{equation}
    \mu_i(\Gnat) \;=\; \mathbb{E}[p(i[V])],
    \label{eq:mu}
\end{equation}
and its companion, the spread of difficulty across variants, is
\begin{equation}
    \sigma^2 \;=\; \mathrm{Var}(p(i[V])).
    \label{eq:sigma}
\end{equation}
Here $\mu_i(\Gnat)$ summarizes \emph{average} performance on variants, while $\sigma^2$
measures \emph{consistency} --- how much the agent's success probability swings across semantically identical inputs. Degradation $\Delta_i(\Gnat) = \mu_i(\Gid) - \mu_i(\Gnat)$ is estimated by $\hat\Delta_i(\Gnat) = \hat\mu_i(\Gid) - \hat\mu_i(\Gnat)$.

\paragraph{Data.} We draw $N$ variants $V_1,\dots,V_N$ independently and run the agent $K$ times on each. Let $X_j$ be the number of the $K$ runs on $V_j$ that resolve the instance, and $\hat p_j = X_j/K$ its resolve rate, so that
\begin{equation}
    X_j \mid V_j \;\sim\; \mathrm{Binomial}(K, p(i[V_j])).
    \label{eq:data}
\end{equation}
Our protocol is $N = 20$, $K = 1$.

\paragraph{A useful identity.} One identity relates the within-variant noise to the two estimands and is used repeatedly below. Write
\begin{equation}
    \nu \;=\; \mathbb{E}\big[p(i[V])(1 - p(i[V]))\big]
    \label{eq:nu}
\end{equation}
for the average within-variant Bernoulli variance. Expanding and applying
$\mathbb{E}[p(i[V])^2] = \mathrm{Var}(p(i[V])) + (\mathbb{E}[p(i[V])])^2 = \sigma^2 + \mu_i(\Gnat)^2$,
\begin{equation}
    \begin{split}
    \nu
    &\;=\; \mathbb{E}[p(i[V])] - \mathbb{E}[p(i[V])^2] \\
    &\;=\; \mu_i(\Gnat) - (\sigma^2 + \mu_i(\Gnat)^2) \\
    &\;=\; \mu_i(\Gnat)(1-\mu_i(\Gnat)) - \sigma^2,
    \end{split}
    \label{eq:identity}
\end{equation}
so that
\begin{equation}
    \sigma^2 + \nu \;=\; \mu_i(\Gnat)(1-\mu_i(\Gnat)).
    \label{eq:identity2}
\end{equation}
The total variability in a single perturbed run therefore splits into a
between-variant part $\sigma^2$ and a within-variant part $\nu$.

\subsection{The estimator}
\label{subsec:est_estimator}

The estimate $\hat\mu_i(\Gnat)$ of $\mu_i(\Gnat)$ averages the per-variant rates,
\begin{equation}
    \hat\mu_i(\Gnat) \;=\; \frac{1}{N}\sum_{j=1}^{N} \hat p_j .
    \label{eq:muhat}
\end{equation}
At $K = 1$ we have $\hat p_j = Y_j \in \{0,1\}$ and this collapses to
\begin{equation}
    \hat\mu_i(\Gnat) \;=\; \frac{\#\{\text{variants resolved}\}}{N},
    \label{eq:muhat_k1}
\end{equation}
an ordinary binomial proportion.

\subsection{Why it is unbiased for any $K$, including $K=1$}
\label{subsec:est_unbiased}

The argument conditions on the variant first and averages over variants second,
using the tower property $\mathbb{E}[\,\cdot\,] = \mathbb{E}\big[\mathbb{E}[\,\cdot
\mid V]\big]$.

\paragraph{Step 1: collapse the agent randomness.} Condition on the $j$-th variant
$V_j$ and write $Y_{j1},\dots,Y_{jK}$ for the $K$ runs on it. By
Eq.~\eqref{eq:bernoulli} each has conditional mean $p(i[V_j])$, so
\begin{equation}
\begin{split}
    \mathbb{E}[\hat p_j \mid V_j]
    &\;=\; \mathbb{E}\Big[\tfrac{1}{K}\textstyle\sum_{k=1}^{K} Y_{jk} \,\Big|\, V_j\Big] \\
    &\;=\; \frac{1}{K}\sum_{k=1}^{K} \mathbb{E}[Y_{jk} \mid V_j] \\
    &\;=\; \frac{1}{K} \cdot K\, p(i[V_j])
    \;=\; p(i[V_j]).
\end{split}
    \label{eq:step1}
\end{equation}
The factors of $K$ and $1/K$ cancel, so this holds \emph{regardless of $K$}. Even a
single draw, $K=1$, is already unbiased for its own $p(i[V_j])$.

\paragraph{Step 2: collapse the variant randomness.} Because $V_j$ is itself random,
the conditional mean in Eq.~\eqref{eq:step1} is a random variable, and averaging it
over the variant population gives, by the tower property and Eq.~\eqref{eq:mu},
\begin{equation}
    \begin{split}
    \mathbb{E}[\hat p_j]
    &\;=\; \mathbb{E}\big[\, \mathbb{E}[\hat p_j \mid V_j] \,\big] \\
    &\;=\; \mathbb{E}\big[\, p(i[V_j]) \,\big]
    \;=\; \mu_i(\Gnat) .
    \end{split}
    \label{eq:step2}
\end{equation}

\paragraph{Step 3: average over variants.} By linearity of expectation, applied to
Eq.~\eqref{eq:muhat},
\begin{equation}
    \begin{split}
    \mathbb{E}[\hat\mu_i(\Gnat)]
    &\;=\; \frac{1}{N}\sum_{j=1}^{N} \mathbb{E}[\hat p_j] \\
    &\;=\; \frac{1}{N} \cdot N\,\mu_i(\Gnat)
    \;=\; \mu_i(\Gnat) .
    \end{split}
    \label{eq:step3}
\end{equation}

\paragraph{Consequence.} The parameter $K$ never appears in
Eqs.~\eqref{eq:step1}--\eqref{eq:step3}. Repetition is a variance-reduction
operation. It tightens each $\hat p_j$ around the center
$p(i[V_j])$, but that center was already correct, so tightening cannot remove a bias that was never there. One run per variant is exactly as unbiased as a thousand.

\subsection{Variance, and why $K=1$ is the efficient allocation}
\label{subsec:est_variance}

Unbiasedness alone does not justify $K = 1$. An unbiased estimator can still be noisy. We now show that $K=1$ is not only admissible but optimal.

\paragraph{Step 1: variance of one per-variant rate.} Decompose $\mathrm{Var}(\hat
p_j)$ with the law of total variance,
\begin{equation}
    \begin{split}
    \mathrm{Var}(\hat p_j)
    &\;=\; \mathrm{Var}\big(\mathbb{E}[\hat p_j \mid V_j]\big) \\
    &\quad\;+\; \mathbb{E}\big[\mathrm{Var}(\hat p_j \mid V_j)\big].
    \end{split}
    \label{eq:ltv}
\end{equation}
The first term is $\mathrm{Var}(p(i[V_j])) = \sigma^2$ by Eqs.~\eqref{eq:step1}
and~\eqref{eq:sigma}. For the second, $\hat p_j = X_j/K$ with $X_j \mid V_j \sim
\mathrm{Binomial}(K, p(i[V_j]))$, so $\mathrm{Var}(\hat p_j \mid V_j) =
p(i[V_j])(1-p(i[V_j]))/K$, whose expectation over variants is $\nu/K$ by
Eq.~\eqref{eq:nu}. Hence
\begin{equation}
    \mathrm{Var}(\hat p_j) \;=\; \sigma^2 + \frac{\nu}{K}.
    \label{eq:varpj}
\end{equation}

\paragraph{Step 2: variance of the estimator.} The variants are drawn independently
and each is run in an isolated environment, so the $\hat p_j$ are i.i.d.\ and
\begin{equation}
    \begin{split}
    \mathrm{Var}(\hat\mu_i(\Gnat))
    &\;=\; \frac{1}{N^2}\sum_{j=1}^{N}\mathrm{Var}(\hat p_j) \\
    &\;=\; \frac{\sigma^2}{N} + \frac{\nu}{NK}
    \;=\; \frac{\sigma^2}{N} + \frac{\nu}{R},
    \end{split}
    \label{eq:varmu}
\end{equation}
writing $R = NK$ for the total number of agent runs.  The
within-variant noise term $\nu/R$ depends only on the \emph{total} budget $R$; how those runs are distributed across variants is irrelevant to it. The between-variant
term $\sigma^2/N$ depends only on the \emph{number of distinct variants} $N$, and $K$ does not appear in it at all.

\paragraph{Step 3: optimize under a fixed budget.} Assume a fixed run budget $R$ and
substitute $N = R/K$ into Eq.~\eqref{eq:varmu}:
\begin{equation}
    \mathrm{Var}(\hat\mu_i(\Gnat))
    \;=\; \frac{K\sigma^2}{R} + \frac{\nu}{R}
    \;=\; \frac{K\sigma^2 + \nu}{R}.
    \label{eq:budget}
\end{equation}
This is strictly increasing in $K$ whenever $\sigma^2 > 0$, and is therefore minimized at $K = 1$. Under a fixed number of agent runs, the variance-optimal design
for estimating the mean is to spend every run on a \emph{new} variant and run the agent once.

\paragraph{Step 4: the $K=1$ specialization.} At $K = 1$ the estimator's variance
simplifies. Substituting the identity~\eqref{eq:identity2} into
Eq.~\eqref{eq:budget} with $K=1$ and $R=N$,
\begin{equation}
    \begin{split}
    \mathrm{Var}(\hat\mu_i(\Gnat))
    &\;=\; \frac{\sigma^2 + \nu}{N} \\
    &\;=\; \frac{\mu_i(\Gnat)(1-\mu_i(\Gnat))}{N}.
    \end{split}
    \label{eq:k1var}
\end{equation}
This has a direct interpretation. Each $Y_j$ is \emph{marginally}
$\mathrm{Bernoulli}(\mu_i(\Gnat))$  and
the $N$ trials are independent, since each uses a distinct variant and no two share a draw. So $\hat\mu_i(\Gnat)$ is a textbook binomial proportion with standard error $\sqrt{\hat\mu_i(\Gnat)(1-\hat\mu_i(\Gnat))/N}$, and $\hat\mu_i(\Gnat)$ may be compared to $\hat\mu_i(\Gid)$ with a standard two-proportion procedure. No clustering correction is needed, because every trial uses a different variant.

\subsection{Probabilistic guarantees}
\label{subsec:est_guarantees}
Because variants are independent draws from an explicitly specified distribution $D_i(g)$ and each run on a variant is an independent Bernoulli trial, our estimates come with finite-sample guarantees of the kind statistical model checking provides~\cite{younes2002probabilistic, legay2010statistical}. With $K = 1$, $\hat\mu_i(g)$ is the mean of $N$ independent $\{0,1\}$ outcomes with expectation $\mu_i(g)$, so by Hoeffding's inequality, with probability at least $1-\delta$,
\[
\mu_i(g) \;\ge\; \hat\mu_i(g) - \sqrt{\ln(1/\delta)/(2N)} .
\]
The same argument applied to the $2N$ independent runs behind $\hat\Delta_i(g) = \hat\mu_i(\Gid) - \hat\mu_i(g)$, each of which moves $\hat\Delta_i(g)$ by at most $1/N$, gives $\Delta_i(g) \le \hat\Delta_i(g) + \sqrt{\ln(1/\delta)\,/\,N}$, and applied to the $2N|\mathcal{I}|$ runs behind $\bar\Delta$ gives a margin of $\sqrt{\ln(1/\delta)/(N|\mathcal{I}|)}$ for the mean degradation. Exact binomial (Clopper--Pearson) bounds are tighter and can replace Hoeffding's. At $N = 20$ and $\delta = 0.05$ the margins are about $0.27$ for $\mu_i(g)$, $0.39$ for $\Delta_i(g)$, and $0.07$ to $0.08$ for $\bar\Delta$ on our instance sets, which is why our claims rest on configuration-level aggregates rather than on individual instances. Sequential tests~\cite{younes2002probabilistic} could stop sampling once a robustness threshold is confirmed or refuted.

These guarantees are conditional. They hold relative to $D_i(g)$, not to the worst case over all equivalent variants or to code an agent may meet in practice; for the fixed instance set $\mathcal{I}$, not for a benchmark viewed as a population; for the keyword map $\Kmap$ extracted for the instance and model; and, since we keep only variants that pass the validation check (Section~\ref{subsec:spt_validation}), for $D_i(g)$ conditioned on passing it.

\paragraph{Relation to the intervals we report.}
The bounds above are exact: they hold at every sample size, and Hoeffding's needs only that each outcome lies in $[0,1]$. Clopper--Pearson intervals are also exact but use the binomial distribution, so they are tighter for a single proportion; they do not extend as simply to a difference of proportions. Section~\ref{sec:experiments} instead reports intervals that are approximate. Newcombe intervals for $\hat\Delta_i(\Gnat)$ combine Wilson score intervals, which invert a normal approximation; their coverage is not guaranteed but stays close to nominal in simulation studies at small samples~\cite{fagerland2015recommended}. Bootstrap percentile intervals for $\bar\Delta$ reach nominal coverage only as the number of runs grows; with 20 runs per condition, instances whose runs all succeed or all fail contribute no resampling variability, so these intervals may be somewhat too narrow. The approximate intervals are much tighter than the exact bounds: the exact margin for $\bar\Delta$ (about 7--8 points) exceeds the largest mean degradation we observe (6.7 points), so an exact Hoeffding bound would not establish that degradation is positive in any configuration. Our claims of significant degradation therefore rest on the approximate intervals. Exact procedures with more power exist: variance-adaptive bounds such as empirical Bernstein bounds, and permutation tests of the null hypothesis that perturbation has no effect, which are exact for any sample size.

\section{Additional Experiment Results}
\label{app:exp_results}

\begin{figure*}[!t]
    \centering
    \includegraphics[width=\columnwidth]{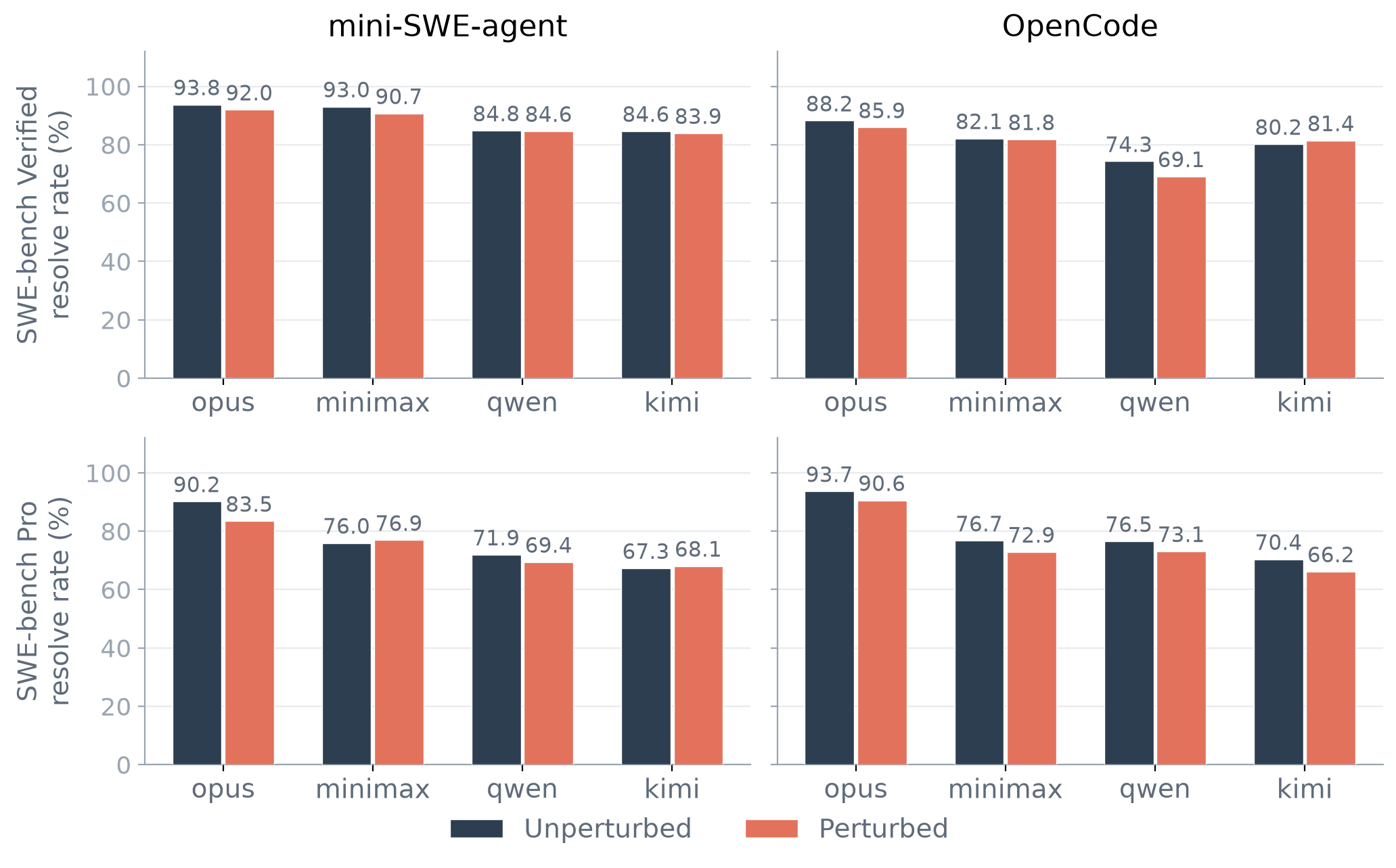}
    \caption{Mean resolve rate (\%) on unperturbed and perturbed runs for each model under each scaffold, on SWE-bench Verified (top) and SWE-bench Pro (bottom).}
    \label{fig:resolve_rate}
\end{figure*}

\paragraph{Resolve rates fall, but capability rankings survive.}
Figure~\ref{fig:resolve_rate} shows the mean resolve rate on unperturbed and perturbed runs for each model under each scaffold, separately on SWE-bench Verified and SWE-bench Pro. In 13 of the 16 scaffold-model-benchmark combinations the agent resolves fewer issues after perturbation.
The three remaining combinations move slightly in the other direction within run-to-run noise. Perturbation, however, barely disturbs the models' \emph{capability} ordering. In three of the four scaffold-benchmark panels, the ranking of models by resolve rate is identical before and after perturbation. In the fourth (OpenCode on SWE-Pro), the only change is a swap between MiniMax and Qwen, which are separated by just 0.2 points at baseline. This shows that a leaderboard-style comparison would not register the impact of SPTs. The ranking that perturbation does scramble is the ranking by \emph{robustness}, as we show next.

\subsection{Effort Overhead over All Runs}
\label{app:effort_all_runs}

\begin{figure*}[!t]
    \centering
    \begin{minipage}[t]{0.48\textwidth}
        \centering
        \includegraphics[width=\linewidth]{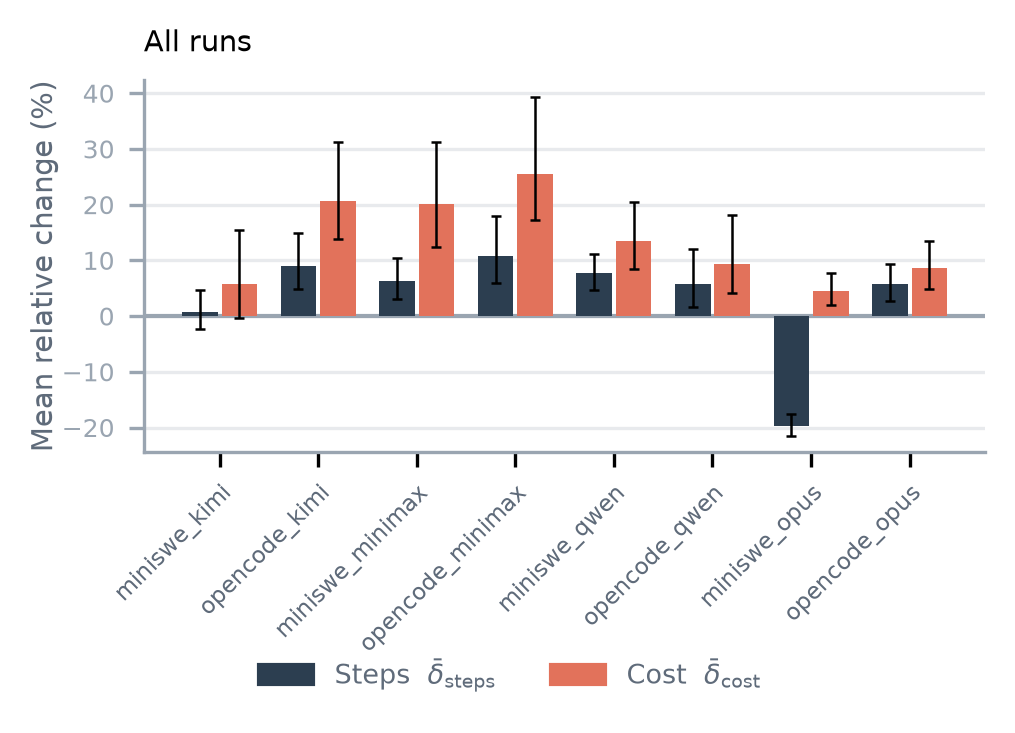}
    \end{minipage}\hfill
    \begin{minipage}[t]{0.48\textwidth}
        \centering
        \includegraphics[width=\linewidth]{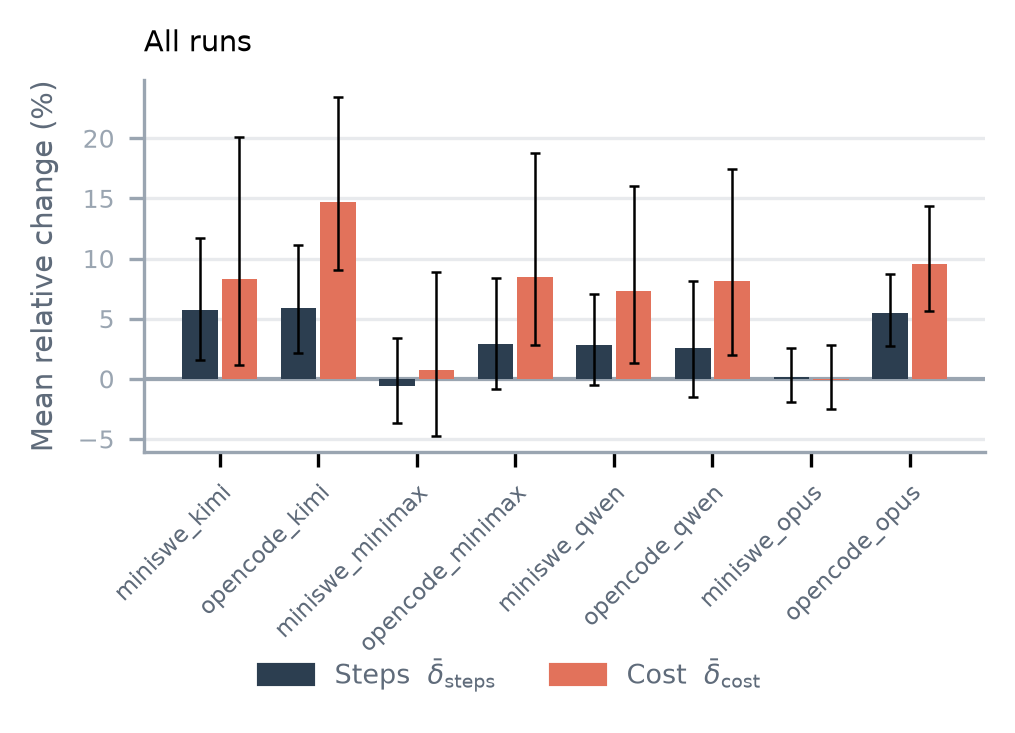}
    \end{minipage}
    \caption{Mean relative change in agent steps ($\bar{\delta}_{\text{step}}$) and cost ($\bar{\delta}_{\text{cost}}$) after perturbation over all runs, for each scaffold--model configuration on the 28 SWE-bench Verified instances (left) and the 26 SWE-bench Pro instances (right). Positive values mean perturbed runs take more steps, or cost more, than unperturbed runs on the same instance. Error bars are fixed-population 95\% bootstrap confidence intervals. }
    \label{fig:cost_step_overhead_all}
\end{figure*}

Figure~\ref{fig:cost_step_overhead_all} repeats the RQ2 effort analysis of the main text over \emph{all} runs, rather than only the runs the agent resolved. The trends match the resolved-only figure. On SWE-bench Verified cost rises in all eight configurations, from 4.5\% (mini-SWE agent with Opus) to 25.6\% (OpenCode with MiniMax), with the interval excluding zero in seven of the eight. On SWE-bench Pro cost rises in seven of the eight, by up to 14.7\% (OpenCode with Kimi), with six intervals excluding zero. The one exception, mini-SWE agent with Opus, is flat at $-0.1\%$. Step counts move less than cost in both benchmarks, and Opus under mini-SWE agent on Verified again takes 19.7\% fewer steps while still costing more. Because the two views agree, the overhead cannot be attributed to long failing runs, and we report the resolved-only view in the main text as the more conservative one.

\subsection{Opus Compresses Its Trajectory}
\label{app:opus_trajectory}
Opus under mini-SWE agent on SWE-bench Verified responds to perturbation by taking \emph{fewer} steps on every instance (mean $-19.8\%$, range $-6\%$ to $-39\%$). Its output tokens per step more than double (median increment $+101\%$) while input tokens per step fall by 17\%, so cost rises on every instance: Opus emits fewer, denser actions, for a net $+30.4\%$ cost per step and $+4.0\%$ total cost. The behavior is benchmark-specific: on SWE-bench Pro the step reduction is only $-1.29\%$.

\subsection{Per-Instance Degradation Plots}
\label{app:per_instance_degradation}

\begin{figure*}[!t]
    \centering
    \includegraphics[width=\textwidth]{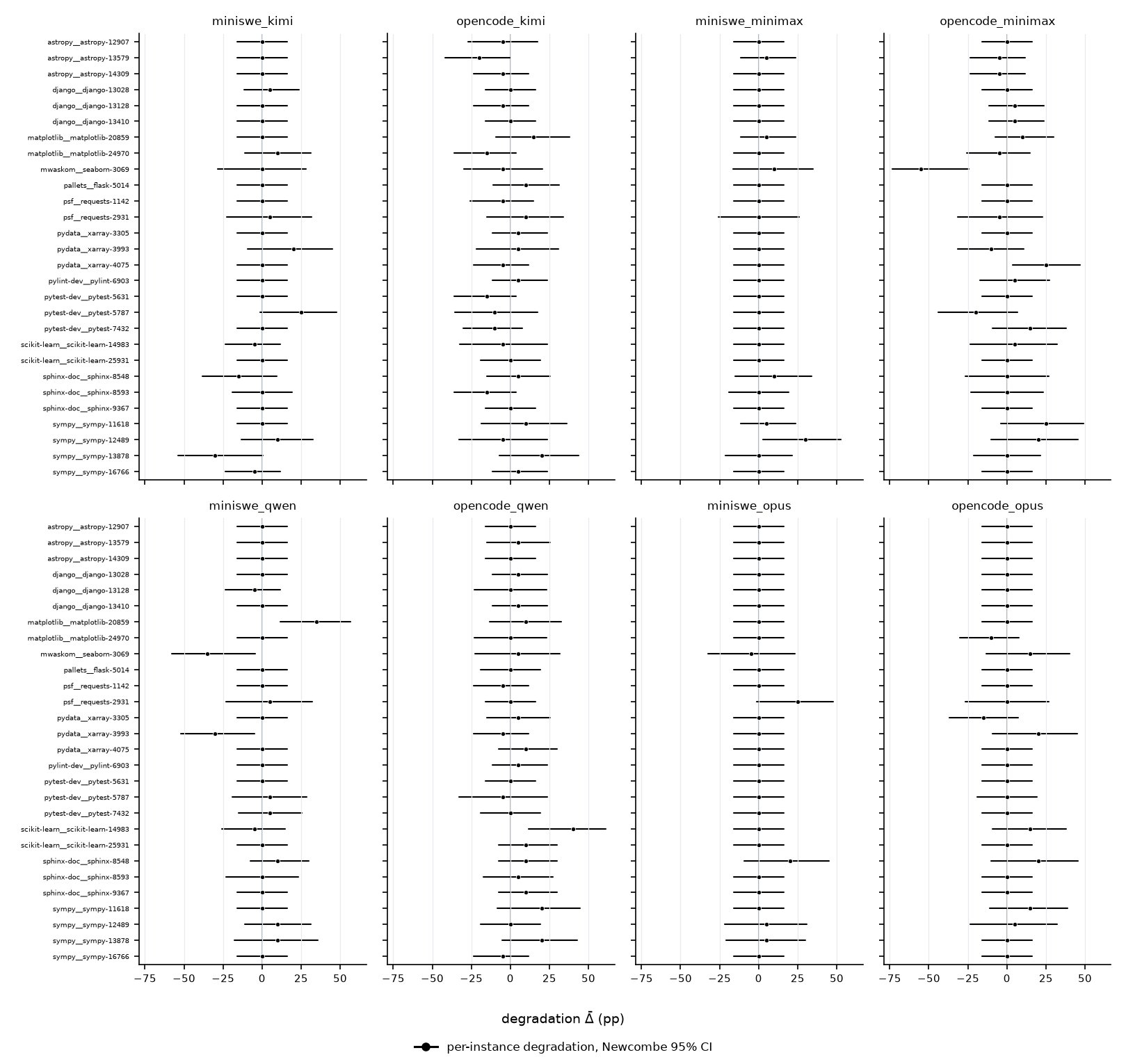}
    \caption{Per-instance degradation $\hat\Delta_i(\Gnat)$ for the 28 SWE-bench Verified instances under each configuration. Error bars are Newcombe 95\% confidence intervals for the difference between the unperturbed and perturbed resolve proportions ($n{=}20$ runs per condition).}
    \label{fig:per_instance_degradation_swebench}
\end{figure*}

\begin{figure*}[!t]
    \centering
    \includegraphics[width=\textwidth]{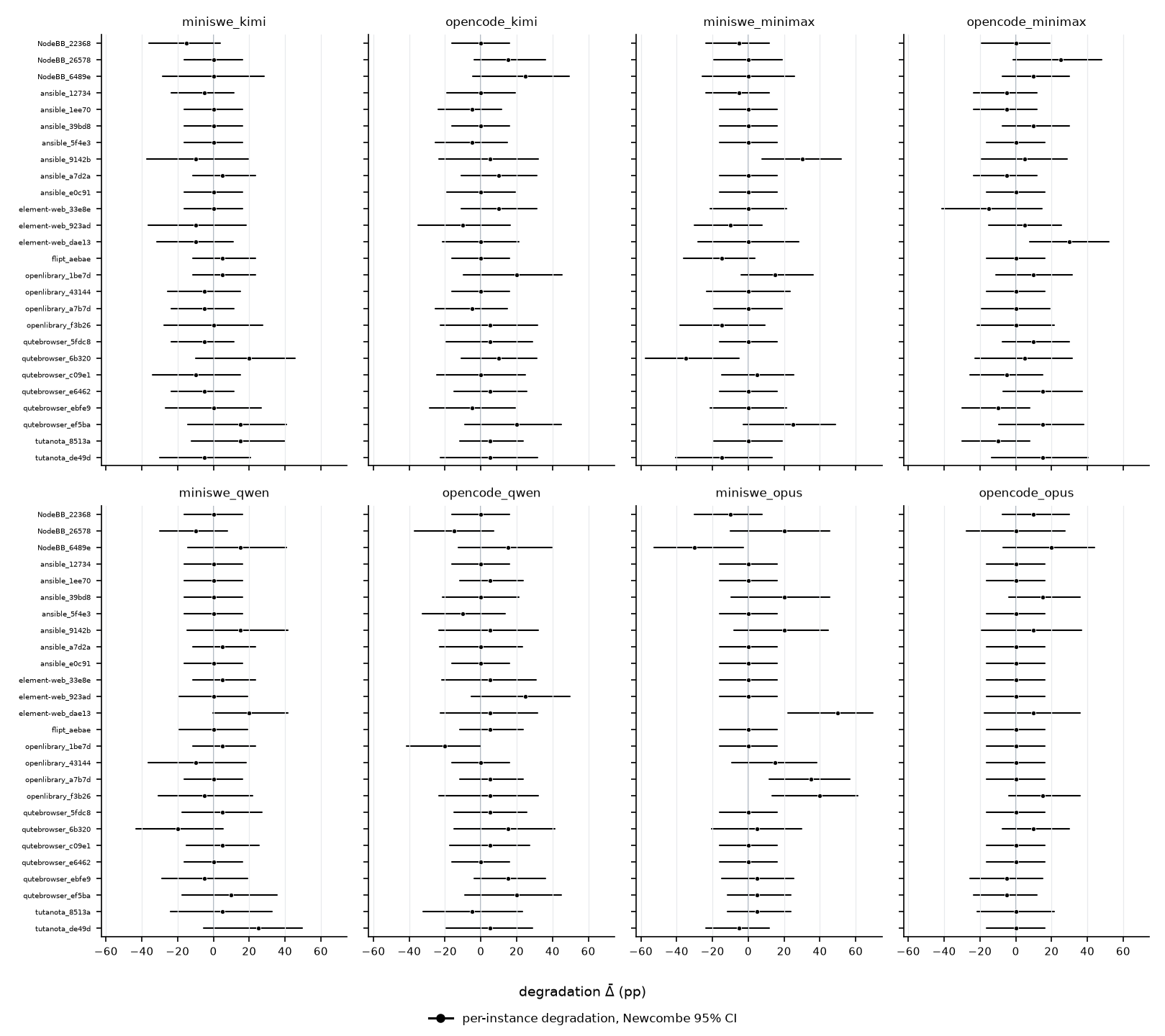}
    \caption{Per-instance degradation $\hat\Delta_i(\Gnat)$ for the 26 SWE-bench Pro instances under each configuration. Error bars are Newcombe 95\% confidence intervals for the difference between the unperturbed and perturbed resolve proportions ($n{=}20$ runs per condition).}
    \label{fig:per_instance_degradation_swepro}
\end{figure*}

Figures~\ref{fig:per_instance_degradation_swebench} and~\ref{fig:per_instance_degradation_swepro} give the per-instance degradation $\hat\Delta_i(\Gnat)$ with its Newcombe 95\% interval for every instance-configuration cell. This section reports the underlying numbers for the concentration claim of the RQ1 results.

With 20 runs per condition, a Newcombe interval is wide by construction: even an instance resolved in every run of both conditions carries a $\pm 16$-point interval. The interval excludes zero in only 7 of the 224 cells on SWE-bench Verified and 7 of the 208 on SWE-bench Pro. Most intervals are therefore too wide to establish degradation for a specific instance, although their mass is shifted toward positive values. Point estimates are positive in 63 cells against 36 negative on SWE-bench Verified, and 79 against 44 on SWE-bench Pro.

The individually significant cells concentrate in the tail. On SWE-bench Verified, \texttt{scikit-learn\_\_scikit-learn-14983} loses 40 points under OpenCode with Qwen, \texttt{matplotlib\_\_matplotlib-20859} loses 35 under mini-SWE agent with Qwen, and \texttt{sympy\_\_sympy-12489} loses 30 under mini-SWE agent with MiniMax. On SWE-bench Pro, \texttt{element-web\_dae13} loses 50 points, \texttt{openlibrary\_f3b26} 40, and \texttt{openlibrary\_a7b7d} 35, all under mini-SWE agent with Opus. These three Pro cells alone account for roughly two thirds of that configuration's 6.7-point mean degradation. Across configurations, \texttt{qutebrowser\_ef5ba} and \texttt{ansible\_9142b} degrade in seven of eight, and \texttt{sympy\_\_sympy-11618} in five while never improving.

Single cells also swing the other way: \texttt{mwaskom\_\_seaborn-3069} improves by 55 points under OpenCode with MiniMax and by 35 under mini-SWE agent with Qwen, both with intervals excluding zero. With a single \texttt{mwaskom\_\_seaborn} instance in the sample, such swings should not be read as a repository-level effect.

\section{Qualitative Analysis Details}
\label{app:qualitative_analysis}

\subsection{Instance Selection Procedure}
\label{subsec:appendix_selection_procedure}

12 of the 20 trajectories are manually picked by evaluating characteristics like high variance (3), high degradation (3), large increase in step count (1) or file coverage (2) or cost (1), or cases with 1/20 variant failures (2). 
The remaining 8 are picked by passing variants that the agent had failed to resolve through an LLM prompt designed to take the SPT log as well as a condensed version of the agent trajectory as input, and predict possible SPT impact based on the SPTs that were applied, intersection of files touched by both the agent and the SPTs, and specific snippets from agent thoughts or tool outputs that have SPT markers. This is done for 115 failing trajectories across all 16 experiments. The output is then manually evaluated to shortlist 8 trajectories for final human review. 

\subsection{AI-Assisted Trajectory Analysis}
\label{subsec:appendix_trajectory_analysis}

The same prompt also extracts a sequence of events from the trajectory, citing snippets from the raw trajectory for each, while also classifying each event as one of the following: Localization (exploring the repo to find relevant code), Debugging (reasoning, specifically based on observed outputs), Planning (forming a plan for patching), Patching (Implementation of changes), Validation (Testing of changes), Recovery (Backtracking on implemented changes or proposed plan) or General (additional events that don't come under any of the 6 aforementioned types, for example, patch submission). A final HTML is generated, containing the raw trajectory, a graph with the extracted events, including snippets from the actual trajectory, a record of files touched by both the agent and the SPTs, and script based results like git statements. Human evaluators then use this HTML to get a comprehensive view of the trajectory while analyzing. This is strictly an assistive method, and all qualitative examples in the paper have involved human review of all relevant portions of the raw trajectories. 

\subsection{Additional Qualitative Insights}
\label{subsec:appendix_more_qual_insights}

\paragraph{Variability in SPT detection.}
There is high variance in whether SPTs are detected at all even in cases where the agent repeatedly accesses mutated code. While some SPTs like double negation are easier to miss, SPTs which have been detected in other cases, like if true wrapper, are ignored in cases like Opus on Psf-Requests with mini-SWE agent. In this particular case, even though the agent accesses mutated code multiple times, it never acknowledges any SPT presence. Moreover, consistent with the qualitative results, SPT impact appears to be variable across instances, with more impact seen in repos like qutebrowser, openlibrary, ansible, pytest and sympy as compared to repos like nodeBB or astropy, based on qualitative reviews.

\paragraph{Additional examples of agent response to SPTs.}
There are multiple other examples of how agents choose to respond to SPTs. In Qwen on Matplotlib with OpenCode, after detecting SPTs, the agent decides to reset the repository. However, it claims to have reset it multiple times without actually doing so, and only later ends up proceeding with it. In another case, Opus on Qutebrowser with mini-SWE agent, the agent notices complicated booleans introduced by the if true wrapper in the relevant code, and spends additional time in resolving the booleans into final true/false values. It later also "fixes" them in the final patch, thereby investing significant effort in interacting with the SPTs. In contrast, MiniMax with mini-SWE agent on Qutebrowser chooses to avoid core files that have been transformed by SPTs, and instead tries to get information from other sources in the repository, subsequently increasing the time taken to localize the issue as well as file coverage.

\paragraph{Input prompt impacts agent jaggedness.}
We observe that issue description quality and nature of issue may impact agent jaggedness. We analyse Qwen, Kimi and Opus on the same NodeBB issue which exhibits high variance across various model and scaffold pairs. The issue description contains two conflicting statements --- editing the code to satisfy issue requirements causes one of the unit tests to break, and the description explicitly states not to alter any unit tests. This leads to all three agents falling into a loop of patching, validating, and debugging, while going back to the issue description multiple times, which substantially adds to both the step count and cost. In a passing trajectory for the same issue, the agent comes across the same conflict and fails the same test, but doesn't backtrack on the code and therefore ends up passing. So the conflicting nature of the issue instructions greatly contributes to jaggedness in agent performance, as it is left to its own judgement to resolve the conflict. 

\paragraph{Robustness of test suite formation varies. }
For Astropy, Kimi uses a three-level test suite, complete with custom tests for basic functionality, project test suite, and a dedicated edge cases script. However, in Pytest and Qutebrowser, the validation framework is brittle, with insufficient debugging and incorrect failure attribution to SPTs in the former and shifting test cases during debugging in the latter. Similarly, for MiniMax, validation setup is comprehensive in Sympy, but has incorrect assertions in custom test scripts in Qutebrowser.

\section{Further Related Work: Robustness of LLMs}
\label{app:related_llm}

LLMs are fragile to semantics-preserving transformations of general natural-language inputs. Prompt formatting changes alone (spacing, delimiters, option ordering) cause up to 76-point accuracy swings~\cite{sclar2024quantifying}, and systematic perturbation benchmarks confirm this across character-, word-, and sentence-level attacks~\cite{zhu2023promptbench, vega2023bypassing}.

The phenomenon extends to structured problem solving. GSM-Symbolic~\cite{mirzadeh2024gsmsymbolic} and GSM-Plus~\cite{li2024gsmplus} create variants of math benchmarks by changing surface features (names, numbers, phrasing) while preserving problem structure, observing up to 10\% performance drops. Similarly, reordering multiple-choice options~\cite{zheng2024robust, chaudhary2024decoding} or generating structurally equivalent reasoning problems~\cite{zhu2024dyval} significantly shifts model rankings. Formal certification of LLM robustness has also been proposed~\cite{chaudhary2025certifying, wang2026catastrophic,chaudhary2025lumos}.

\citet{geirhos2020shortcut} frame these failures as \emph{shortcut learning}, where networks exploit spurious surface statistics rather than learning the intended decision rule. \citet{dziri2023faith} provide evidence that transformers linearize compositional reasoning, offering a mechanistic explanation for why surface changes can derail multi-step tasks.

Our findings echo this pattern in the code domain. We show coding agents are sensitive to how the surrounding code is structured. A further dimension absent from prior work is the scaffold interaction. We find that robustness rankings can change when the same model is embedded in a different agent architecture. Robustness certification for agentic systems has also been studied for simple, non-code tool-selection~\cite{yeon2025quantifying}.

\end{document}